\PassOptionsToPackage{table}{xcolor}                                                  
\documentclass[table,10pt,twocolumn,letterpaper]{article}

\usepackage{cvpr}              

\newcommand{\rz}[1]{{\color{black}#1}}
\newcommand{\rzz}[1]{{\color{black}#1}}

\newcommand{\rqw}[1]{\textcolor{black}{#1}}

\newcommand{\iccv}[1]{\textcolor{black}{#1}}

\usepackage{multirow}
\usepackage{algorithm}
\usepackage[noend]{algpseudocode}
\newcommand{\argmax}{\operatornamewithlimits{argmax}}

\definecolor{cvprblue}{rgb}{0.21,0.49,0.74}
\usepackage[pagebackref,breaklinks,colorlinks,allcolors=cvprblue]{hyperref}
\usepackage{comment}
\usepackage{ragged2e}
\usepackage[T1]{fontenc}
\usepackage{courier}
\usepackage{inconsolata}
\usepackage[symbol]{footmisc}
\def\paperID{6645} 
\def\confName{ICCV}
\def\confYear{2025}

\title{RESAnything: Attribute Prompting for Arbitrary Referring Segmentation}

\newcommand{\ourmethod}{RESAnything}
\newcommand{\ourdataset}{ABO-Image-ARES}
\author{Ruiqi Wang \quad Hao Zhang\\
Simon Fraser University
}

\begin{document}
\makeatletter
\let\@oldmaketitle\@maketitle
\renewcommand{\@maketitle}{\@oldmaketitle
  \includegraphics[width=\linewidth]
    {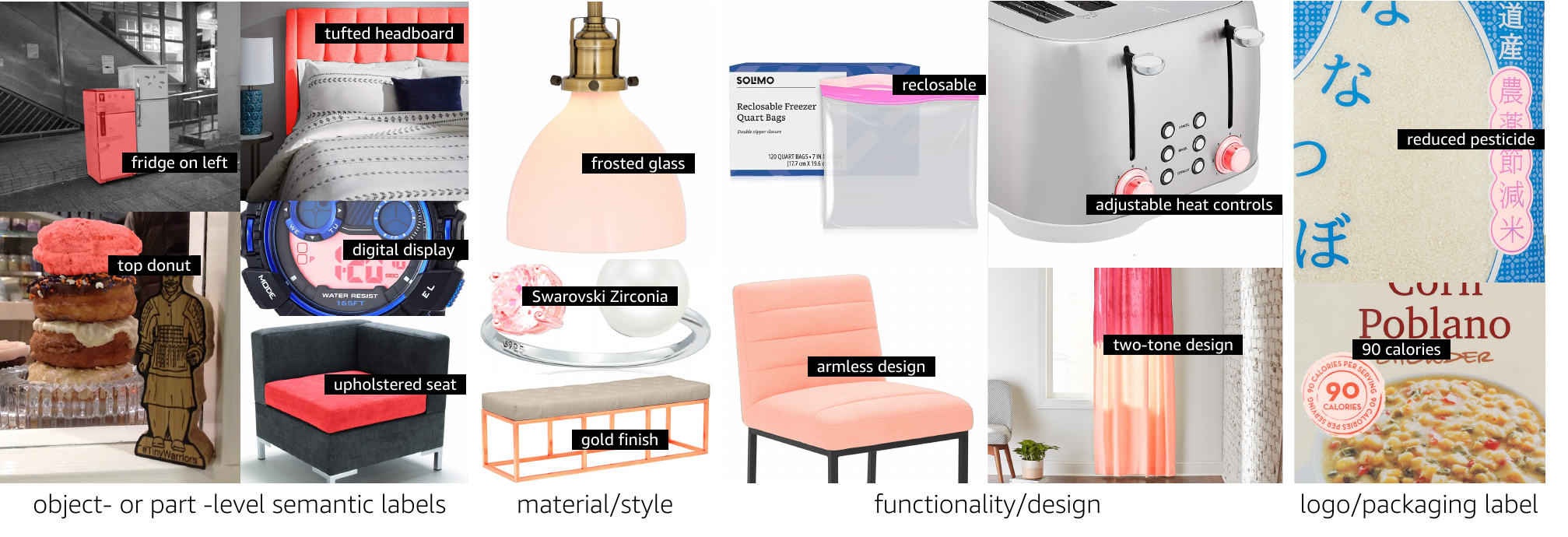}
    \vspace{-1em}
    \captionof{figure}{\rz{\textbf{Open-vocabulary and zero-shot referring expression segmentation with \ourmethod.} Our method produces accurate object or part masks from general- and free-form text expressions including, from left to right: object or part semantic label, material/style properties, function/design descriptions, or logos and packaging labels in textual or other graphical in an image. For visualization purposes, we overlay segmentation regions with red color in each example.}}
    \label{fig:teaser}\bigskip}
\makeatother
\maketitle
\begin{abstract}
\rz{
We present an {\em open-vocabulary\/} and {\em zero-shot\/} method for {\em arbitrary\/} referring expression segmentation (RES), 
targeting more general input expressions than those handled by prior works. Specifically, our inputs encompass 
both object- and {\em part-level\/} labels as well as {\em implicit\/} references pointing to {\em properties\/} or {\em qualities\/} 
of object/part function, design, style, material, etc.
Our model, coined \ourmethod, leverages {\em Chain-of-Thoughts\/} (CoT) reasoning, where the key idea is {\em attribute prompting\/}. We generate detailed descriptions of object/part attributes 
including shape, color, and location for potential segment proposals through systematic prompting of a large language model (LLM), 
where the proposals are produced by a foundational image segmentation model. Our approach encourages deep reasoning about 
object/part attributes related to function, style, design, etc., to handle implicit queries without any part annotations for training or fine-tuning.
As the first zero-shot and LLM-based RES method, \ourmethod~achieves superior performance among zero-shot methods on traditional RES benchmarks and significantly outperforms existing methods on challenging scenarios involving implicit queries and complex part-level relations.
We contribute a new benchmark dataset of $\sim$3K carefully curated RES instances to assess part-level, arbitrary RES solutions.
}
\end{abstract}    
\vspace{-2em}
\section{Introduction}
\label{sec:intro}

With rapid developments in Large Multimodal Models (LMMs), visual perception systems have evolved significantly, demonstrating remarkable capabilities in bridging vision and language tasks~\cite{dai2023instructblip, gao2023llama, li2023otter, liu2024visual}. Recent advancements in LMMs have enabled sophisticated understanding of visual content, from object detection to semantic segmentation~\cite{alayrac2022flamingo, chen2023shikra, peng2023kosmos}. 
One of the emerging segmentation tasks that has drawn a great deal of attention lately is the so-called Referring Expression Segmentation (RES) which aims at obtaining a segmentation mask in an image or video that represents an object instance referred to by a natural language expression~\cite{ding2021vision,wang2022cris,yang2022lavt,liu2023gres,yu2023zero,kazemzadeh2014referitgame,you2023ferret,wu2023uniref,chen2025sam4mllm}.

Despite much progress made on RES, two common limitations are often observed.
First, while existing approaches excel at identifying and segmenting objects as whole entities, they often fall short when the input expressions refer to specific object {\em parts\/}. Such situations arise frequently in applications such as eCommerce, where sellers and buyers often promote or review product features referring to specific parts, and in robotics, human-computer interaction, and automated systems, where agents must interact with object parts.
Second, most works to date on RES have focused on referring expressions that contain semantic labels in one way or another. Even the so-called generalized RES (GRES)~\cite{liu2023gres} only extends the expression coverage to an arbitrary number of (including zero) target objects, {\em with labels\/}. On the other hand, object/part references are often {\em implicit\/}, without semantic labels. Such expressions can refer to {\em properties\/} or {\em qualities\/} related to object/part function, design, style, material, or they may appear in textual or other graphical forms as a logo or packaging label; see Fig.~\ref{fig:teaser} for some samples expressions and segmentations.


\begin{figure}[t]
    \centering
    \includegraphics[width=.95\linewidth]{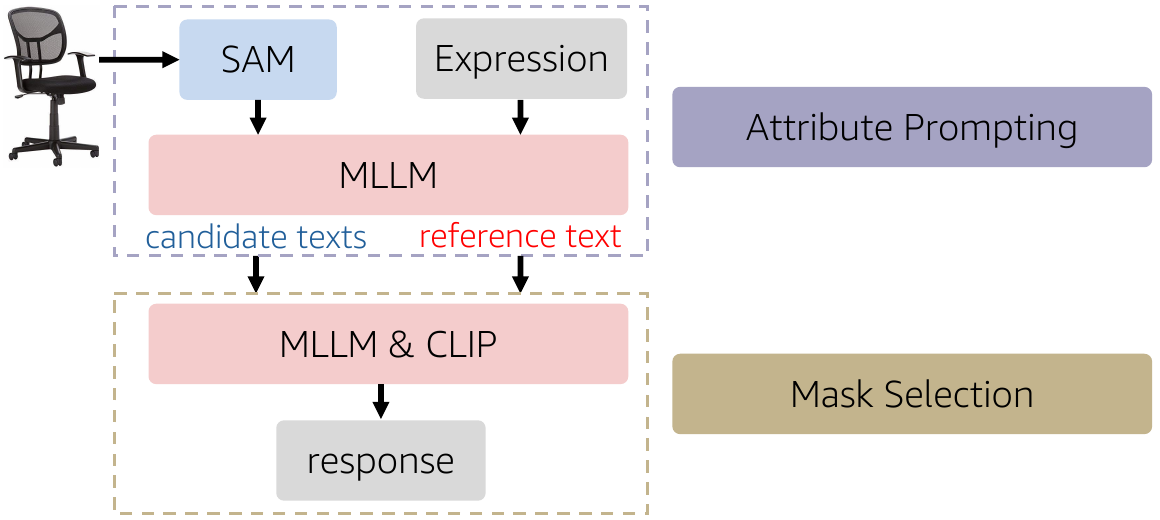}
    \caption{Overview of \ourmethod: a two-stage framework for zero-shot arbitrary RES. The attribute prompting stage generates reference and candidate texts from input image and referring expression using SAM-generated proposals and an MLLM. The mask proposal selection stage leverages MLLM and CLIP to evaluate both candidates and proposals and produce the final response.}
    \label{fig:overview}
    \vspace{-1.5em}
\end{figure}

In this paper, we present an {\em open-vocabulary\/} and \rz{{\em zero-shot\/}} RES method to address both limitations. For lack of a better term, we call our task {\em arbitrary\/} referring segmentation and our  model as {\em \ourmethod\/}. \rz{Our goal is to allow input expressions to be more general than what prior works have been designed to handle, while solving our problem without any training or fine-tuning on specialized datasets.} 
To this end, we leverage the generalization and zero-shot capabilities of modern-day foundational models such as Pixtral~\cite{agrawal2024pixtral} and Claude~\cite{claude-sonnet} as Large Language Models (LLMs and SAM~\cite{kirillov2023segment} for image segmentation.
However, solving the arbitrary RES task demands a deeper understanding of object and part properties, moving beyond traditional object-level and label-centric referencing to more nuanced reasoning for part- and attribute-level perception.


There have been recent works~\cite{lai2024lisa,rasheed2024glamm,lan2024text4seg} on reasoning-based segmentation through active LLM querying. An implicit query text, such as ``the object containing the most Vitamin C,'' is first analyzed by a text LLM and then referenced to the ``orange'' object in the provided image. Nonetheless, such methods often fall short when \rz{the implicit connections between object/part properties (e.g., functional or stylistic ones) and their visual manifestations} are cascadedly hidden. Even advanced LLMs, with their sophisticated reasoning capability, struggle to ground their understanding without explicit supervision at the part or attribute level. 
Additionally, existing methods, \rz{e.g., LISA~\cite{lai2024lisa}, typically rely on fine-tuning on specially prepared or curated datasets --- they are {\em not zero-shot\/}.} 

\rz{Our model for arbitrary RES is {\em training-free\/}.} It leverages {\em Chain-of-Thoughts \/}(CoT) for \rqw{comprehensive} part-level understanding. \rz{Our key idea is} {\em attribute prompting\/}, which generates detailed descriptions of \rz{object/part attributes including shape, color, and location} for \rz{potential segment} 
proposals through systematic prompting of LLMs~\cite{agrawal2024pixtral,claude-sonnet}, \rz{where the proposals are produced by a foundational image segmentation model such as SAM~\cite{kirillov2023segment}.} Our approach encourages deep reasoning about \rz{object/part attributes related to function, style, design, etc.,} enabling the system to handle implicit queries without 
\rz{any part annotations for training or fine-tuning.} By bridging abstract descriptions with concrete visual attributes through \rz{a {\em two-stage\/} evaluation framework (attribute prompting + grouping and selection of segment proposals), as illustrated in Fig.~\ref{fig:overview}, \ourmethod~achieves robust performance on both traditional referring expressions and challenging implicit queries for arbitrary RES.}



In summary, our contributions are as follows:
\begin{itemize}
\item \rz{The {\em first zero-shot\/} and {\em LLM-based open-vocabulary RES\/} method, targeting input expressions that are more general than those addressed by prior works.}
\item The novel idea of attribute prompting, as a means for Chain-of-Thoughts (CoT) reasoning, to achieve SOTA performance on both object- and part-level RES tasks.
\item A new dataset, \ourdataset, built upon ABO~\cite{collins2022abo}, offering carefully curated RES instances as a benchmark to assess part-level, arbitrary RES solutions. 

Our dataset consists of 2,989 expression-segment pairs: 1,360 with object/part semantic labels, 742 depicting logos/packaging labels, 502 referring to functions/designs, and finally, 385 covering material/style properties.
\end{itemize}

\vspace{3pt}

We demonstrate by extensive experiments that \ourmethod~achieves \rz{superior\/} performance among zero-shot methods on traditional RES benchmarks \rz{such as RefCOCO, RefCOCO+~\cite{yu2016modeling}, RefCOCOg~\cite{mao2016generation, nagaraja2016modeling}.} \rz{Our method also significantly outperforms existing methods on the recent 
reasoning segmentation dataset ReasonSeg~\cite{lai2024lisa},} as well as RES tasks in challenging scenarios involving implicit queries and complex part-level relationships \rz{such as those from \ourdataset.}
\rzz{With its zero-shot capabilities, the most important practical advantage of our method lies in the improved scalability and generalizability for real-world applications with diverse referring expressions. In contrast, current supervised methods, e.g., LISA~\cite{lai2024lisa} and GLaMM~\cite{rasheed2024glamm}, require substantial training resources, with high data collection and annotation costs by humans. While performing well on vanilla RES benchmarks, they are not as scalable and are limited to scenarios in their training data.}

\section{Related Work}
\label{sec:related-work}



\rz{Recently, multimodal LLMs (MLLMs) has brought the success} of LLMs to image understanding by integrating the visual and linguistic modalities. Example state-of-the-art proprietary models include Claude Sonnet~\cite{claude-sonnet}, Gemini~\cite{google-gemini}, GPT-4 series~\cite{openai-gpt4} etc. Most existing MLLM architectures connect a pre-trained vision encoder to the LLM decoder with a modality connector. For example, Flamingo~\cite{alayrac2022flamingo} proposed the Perceiver Resample to bridge the modality gap, with follow-up works OpenFlamingo~\cite{awadalla2023openflamingo} and Otter~\cite{li2023mimic} particularly developed for effective in-context instruction tuning. InstructBLIP~\cite{dai2023instructblip} built upon the Querying Transformer as in BLIP2~\cite{li2023blip}. The LLaVA models~\cite{liu2024visual,liu2024improved} and Mini-GPT4~\cite{zhu2023minigpt} utilized a lightweight MLP and achieved appealing performances in various MLLM benchmarks. Recent developments include supporting high-resolution image inputs~\cite{xu2024llava,liu2024llava,zhang2024beyond}, optimizing model efficiency~\cite{bai2023qwen,ye2024mplug,zhou2024tinyllava}, and constructing higher-quality datasets~\cite{chen2023sharegpt4v,deitke2024molmo}. 

\subsection{Open-Vocabulary and RES}

RES~\cite{kazemzadeh2014referitgame,nagaraja2016modeling,hu2016segmentation} aims to segment target image regions based on textual descriptions. The core challenge lies in bridging the gap between image and language modalities. Typically, transformer-based text encoders~\cite{devlin2018bert,radford2021learning} are employed to extract textual embeddings, which are then integrated into segmentation architectures through cross-attention or feature alignment~\cite{chng2024mask,shi2018key,wei2023linguistic,ye2019cross,zhang2022coupalign,xiao2024florence} \iccv{to achieve language-aware segmentation~\cite{li2021referring,yang2022lavt,wang2022cris,liu2023polyformer, wang2024unveiling, xiao2024oneref}.}
Recently, SAM~\cite{kirillov2023segment} has introduced text-guided segmentation~\cite{zhang2024evf, liu2025refersam, chen2025sam4mllm}. For instance, Grounding-SAM~\cite{ren2024grounded} leverages bounding boxes returned by Grounding-DINO~\cite{liu2023grounding} to prompt SAM for mask prediction, while Fast-SAM~\cite{zhao2023fast} utilizes CLIP similarity scores~\cite{radford2021learning}  to select the final result from class-agnostic masks generated by SAM. However, the majority of these methods have been primarily designed for object-level segmentation based on explicit semantic expressions.

To address a broader range of segmentation targets and linguistic inputs beyond semantics, methods based on MLLMs have emerged, leveraging the powerful language understanding capabilities inherited from \iccv{LLMs~\cite{zhang2023next,chen2025sam4mllm,lan2024text4seg,you2023ferret,yuan2024osprey,wang2024visionllm,chen2023shikra,peng2023kosmos,zhang2023gpt4roi,xu2024pixel, dai2024simvg, pramanick2024jack}.} One of the pioneering works in this area is LISA~\cite{lai2024lisa}, which enables MLLMs to segment objects by using text embeddings from LLaVA to prompt a SAM~\cite{kirillov2023segment} decoder to predict masks. 
LISA demonstrated promising performance on a new task called Reasoning Segmentation, similar to our Arbitrary Referring Segmentation. 
While improvements over LISA have been developed for extending it to generalized RES~\cite{xia2024gsva,wu2024see} and grounded segmentation~\cite{rasheed2024glamm,ren2024pixellm}, fine-tuning MLLMs on fixed segmentation datasets not only restricts the variety of referring expressions but also weakens the reasoning capability of pre-trained MLLMs. 
In contrast, our method operates in a training-free manner, preserving the complete ability of the MLLM to reason about the input images.

Some methods have demonstrated the feasibility of adopting pre-trained foundation models for RES without additional training~\cite{yu2023zero,karazija2023diffusion,suo2023text,zhou2022extract,sun2024clip}. MaskCLIP obtains pseudo masks by modifying the last attention layer of CLIP~\cite{zhou2022extract}. 
CaR couples CLIP and GradCAM to generate mask proposals, then employs a CLIP classifier to select the final masks, before a mask refinement~\cite{sun2024clip} in post-processing. \iccv{Global-Local CLIP~\cite{yu2023zero} pioneered zero-shot RES
using CLIP to extract visual features.} Our approach follows a similar design, leveraging SAM for proposal generation and MLLMs for mask selection. Although MLLMs already exhibit superior reasoning abilities compared to CLIP, our novel attribute promoting technique further amplifies their inferential capabilities for arbitrary RES.

\subsection{Visual Prompting}

Prompting~\cite{sahoo2024systematic}
has emerged as a powerful technique for adapting pre-trained language models to downstream applications. By incorporating additional hand-crafted instructions, prompt engineering methods effectively facilitate the adaptation process. For instance, Chain-of-Thought (CoT) prompting encourages models to explain their step-by-step reasoning while answering questions~\cite{wei2022chain}. Recently, \iccv{visual prompting~\cite{ yang2023fine, nag2024safari, shtedritski2023does, sun2024vrp}} has been proposed to enhance the adaptation of CLIP for open-vocabulary segmentation by overlaying ovals over segmentation targets~\cite{sun2024clip}. SAM~\cite{kirillov2023segment}, on the other hand, allows users to provide points, boxes, masks as prompts for image segmentation, with the latest version supporting video segmentation~\cite{ravi2024sam2}. Visual prompting has also been applied to MLLMs~\cite{wu2024visual}. Overlaying image regions with bounding boxes, masks, circles, scribbles, and numeric markers has enhanced MLLMs' ability to perform region or pixel-level image understanding~\cite{yang2023dawn,yang2023set,cai2024vipllava}.

\begin{figure*}[t]
    \centering
    \includegraphics[width=\linewidth]{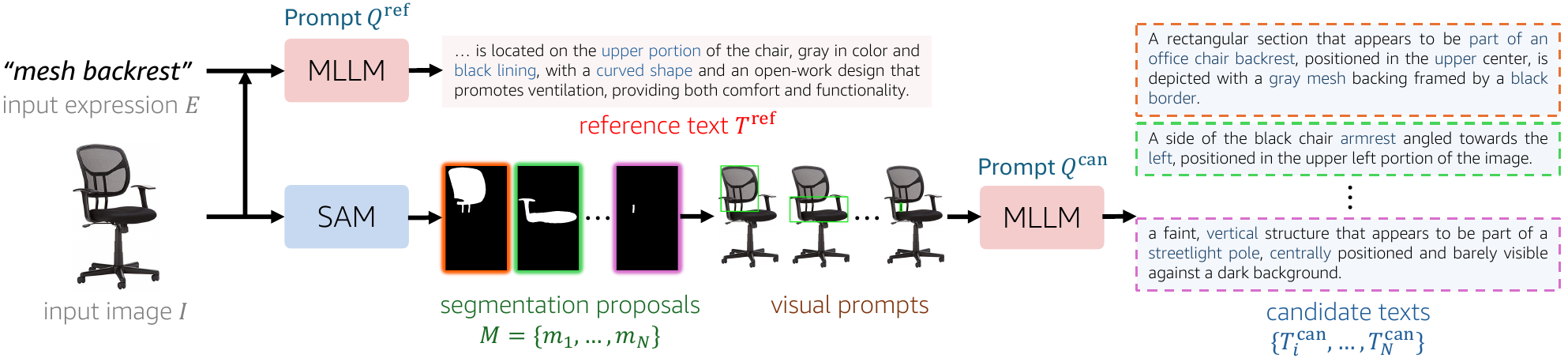}
    \vspace{-20pt}
    \caption{Attribute prompting using SAM and MLLM. Given the input image and referring expression, this stage produces two groups of predictions. The first output, a reference text $T^\text{ref}$, is generated from an MLLM with the text prompt $Q^\text{ref}$. It describes the visual attributes (e.g., color, shape, location) of the target region (``mesh backrest" in this example). The second group is a set of candidate texts $T_i^\text{can}$, generated by an MLLM with the text prompt $Q^\text{can}$ and visual prompts derived from segmentation mask proposals. These texts describe the attributes of their corresponding segmentation region proposals, visualized with the same border color.
    } 
    \label{fig:attribute-prompting}
\end{figure*}


\section{Method}
\label{sec:method}

\paragraph{Problem statement.}
Given an image $I$ and a free-form expression $E$ referring to a potential target region $R$ in $I$, \ourmethod~first processes the image to generate and refine a set of segmentation proposals $M = \{m_1, \ldots, m_N\}$, from which it selects the most appropriate binary segmentation mask $m_i$ representing $R$. The input expression $E$ can be either an explicit referring expression (e.g., semantic label of an object/part) or an implicit expression (e.g., functional or material properties). For targets not directly visible, our method handles two scenarios:
a) Irrelevant queries: indicate that the target does not exist in the image;
b) Invisible targets: infer their location through their functional and spatial relationships, with explanatory reasoning.


A naive approach for applying MLLMs to solve our task would involve prompting the MLLMs to output a score for each segmentation proposal $m_i$, indicating its similarity to the input expression $E$.
However, current MLLMs struggle with directly connecting the text description to the image region.
It is possible to fine-tune a MLLM with many paired samples of texts and mask annotations, however, as mentioned earlier, this incurs significant computational cost during fine-tuning and human effort for data annotation.

\vspace{-1em}

\paragraph{Overview.}
Instead of fine-tuning, we propose a novel approach to facilitate reasoning between text descriptions and visual elements, by systematic ``attribute prompting," which tasks the MLLMs with generating detailed text descriptions of visual properties including shape, color and location.
By doing so, we not only encourages the MLLMs to perform in depth visual reasoning around the target regions, but also circumvents MLLMs weakness in handling image-text pairs, by creating additional intermediate text-text pairs that enable more robust comparison metrics.

Figure~\ref{fig:overview} provides an overview of \ourmethod{}, which consists two main stages:
1) an attribute prompting stage that generates reference text for the target and candidate texts for generated segmentation proposals (Section~\ref{subsec:attribute-prompting}); 2) a proposal selection stage that employs multiple metrics to robustly analyze the relationship between candidate and reference texts and produce the final response (Section~\ref{subsec:grouping-selection}).


\begin{figure}[t]
    \vspace{-1em}
    \centering
    \includegraphics[width=\linewidth]{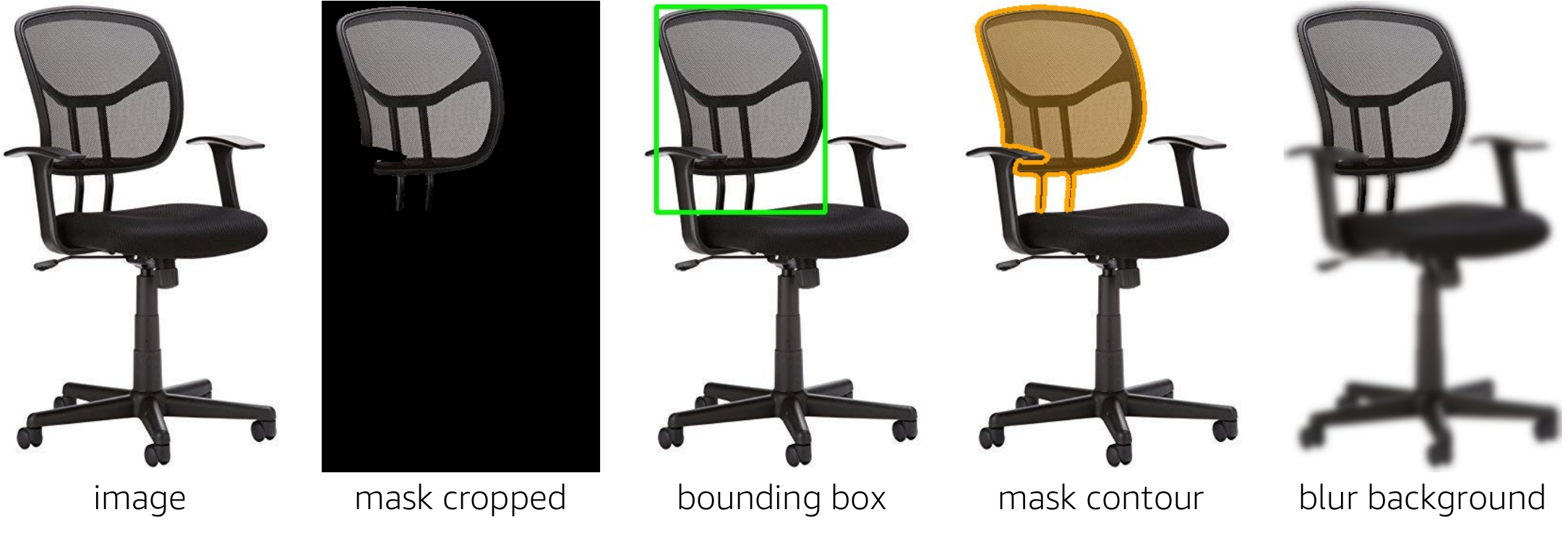}
    \vspace{-20pt}
    \caption{Example of different visual prompts $V_i$ generated from a segmentation proposal $m_i$.}
    \label{fig:visual-prompt}
    \vspace{-10pt}
\end{figure}

\begin{figure*}[t]
    \centering
    \includegraphics[width=.92\linewidth]{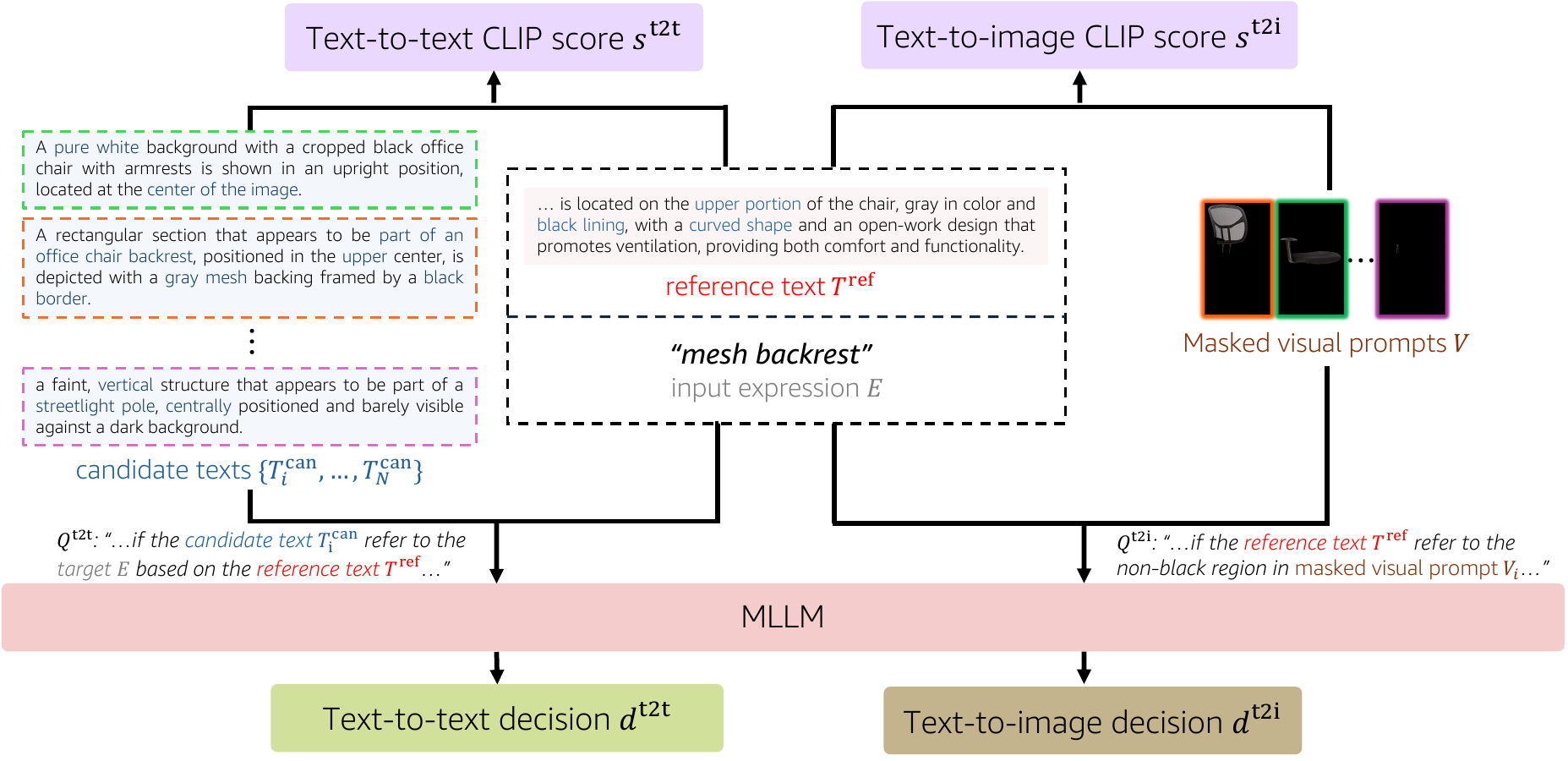}
    \caption{Multi-metric mask proposal selection using MLLM and CLIP. To select the final mask from mask proposals generated by SAM, we introduce four metrics computed across different modalities and models to evaluate the similarity between input expression $E$ and the mask proposals. Specifically, the text-to-text MLLM-based binary decision $d^\text{t2t}$ and CLIP score $s^\text{t2t}$ match reference text to candidate texts. The text-to-image MLLM-based binary decision $d^\text{t2i}$ and CLIP score $s^\text{t2i}$ match reference text to masked visual prompts.}
    \label{fig:voting}
    \vspace{-1.5em}
\end{figure*}

\label{subsec:method-overview}

\subsection{Text Generation via Attribute Prompting}
\label{subsec:attribute-prompting}

To facilitate reasoning between the input expression $E$ and the segmentation proposals $M$, we first apply attribute prompting to generate detailed text descriptions: reference text $T^\text{ref}$, which describes the input expression $E$ in relation to the image $I$, candidate texts $T_{1\ldots N}^\text{can}$, which describe each of the segmentation proposals in a format similar to that of the reference text.
We apply MLLMs to generate these texts, carefully designing the input prompts to encourage the MLLMs to provide description that capture comprehensive object properties and inter-object relationships.

\vspace{-1em}
\paragraph{Reference text generation.}
The reference text $T^\text{ref}$ functions as an extended visual description of the input expression $E$, providing more concrete visual attributes for challenging expressions such part-level semantic labels and functionality/feature-based descriptions.
We task a MLLM to generate the reference text $T^\text{ref} = f_\text{MLLM}(I, E
\;|\;Q^\text{ref})$, with a carefully designed reference text prompt $Q^\text{ref}$ that instructs the MLLM to generate a single sentence with detailed visual attributes, such as shape, color and location, that describe the region $R$ in $I$ targeted by $E$.
For invisible or irrelevant targets, the $T^\text{ref}$ provides a reasoned explanation of why the target cannot be localized. 
We provide the full reference text prompt $Q^\text{ref}$ in the supplementary.
An example is shown in the top part of the Fig~\ref{fig:attribute-prompting}. Given the input ``mesh backrest", the reference text describes its key attributes: ``a \emph{gray curved mesh} backrest with \emph{black lining} located at the \emph{upper portion} of the chair". 

\vspace{-1em}
\paragraph{Candidate text generation.}
The candidate texts $T^\text{can}_1$, $\ldots$, $T^\text{can}_N$ describe the mask proposals $m_1$, $\ldots$, $m_N$ in a format similar to that of the reference text $T^\text{ref}$. 
Without requiring fine-tuning, our method can directly apply off-the-shelf SOTA image segmentation methods to obtain mask proposals.
We adopt SAM~\cite{kirillov2023segment} in this work.
As SAM's raw outputs often contain duplicate or overlapping masks, as well as tiny segments, we configure SAM with sampling points at 0.015\% of total image pixels and filter out segments smaller than 0.1\% of the image area, preventing over-segmentation while maintaining meaningful region proposals. We also filter out duplicate proposals.

Given a mask proposal $m_i$, we generate a corresponding candidate text $T^\text{can}_i = f_\text{MLLM}(V^1_i, V^2_i \ldots V^K_i \;|\; Q^\text{can})$ using an MLLM, where $Q^\text{can}$ is the candidate text prompt that similarly asks for visual attributes such as shape, color and location; and $V^1_i \ldots V^K_i$ are $K$ visual prompts that provide distinct visual representations of the mask proposal $M_i$.
A good visual prompt need to guide the MLLM to focus on the mask region, without removing attribute-related information or adding distractions.
Figure~\ref{fig:visual-prompt} shows a few possible representations for visual prompts:
\textit{image} retains all information of the original image, but does not cover any mask-specific properties;
\textit{mask cropped} highlights the visual attributes of the masked region, but does not suggest the location of the masked region nor its relation with other parts of the image;
in contrast, \textit{bounding box}, \textit{mask contour} and \iccv{\emph{blur background}} provides such relational and locational information, but the bounding box outlines, the mask overlays, \iccv{and blur background} are distractions when it comes to visual properties such as color or shape. 
Using multiple visual prompts, intuitively, alleviate the issues of the respective prompting representation.
In practice, we find using two visual prompts, \textit{bounding box} ($V^b$) and \textit{mask cropped} ($V^m$), is sufficient for our purpose. 
This is consistent with the observations of~\cite{sun2024clip}.
The complete candidate text prompt $Q^\text{can}$ is provided in the supplementary.
Fig~\ref{fig:attribute-prompting}, right part shows examples of generated candidate texts.


\subsection{Multi-metric Mask Proposal Selection}
\label{subsec:grouping-selection}

The generated reference text and candidate texts allow us to assess the similarity between the input expression $E$ and the mask proposals $M$ much more effectively: the reference text $T^\text{ref}$ provides more detailed information than the original expression $E$, thus facilitating in depth text-to-image comparisons; in addition, the candidate texts $T^\text{can}$ enables an additional modality, allowing direct comparisons between two piece of texts.
In this stage, we combine multiple evaluation metrics to perform both text-to-image and text-to-text comparisons to select the mask proposal (or none) that matches the input expression.

\vspace{-1em}
\paragraph{Text-to-text comparison.}
To compare a mask proposal $m_i$ against the input expression $E$, we first evaluate the similarity between the reference text describing $E$, and the candidate text describing $m_i$.
We first use the same MLLM to generate a binary decision 
$d_i^\text{t2t} = f_\text{MLLM}(T^\text{ref}, T_i^\text{can} \;|\; Q^\text{t2t}) \in \{ 0, 1 \}$, 
where $Q^\text{t2t}$ is the text-to-text comparison prompt, as shown in the lower left corner of Figure~\ref{fig:voting}.
The MLLM outputs a yes/no binary decision, as we observed empirically that it often struggles to output consistent scalar scores.
However, there are cases where multiple mask proposals receive a ``yes" response. To disambiguate such cases, we further employ CLIP to generate a scalar similarity score:
$s_i^\text{t2t} = f_\text{CLIP}(T^\text{ref}, T_i^\text{can}) \in [ 0, 1 ]$. 
Although CLIP is generally more error-prone (as we show in the supplementary), its ability to output consistent scalar scores makes it well-suited for further disambiguating among the top candidates filtered by the binary MLLM decision.

\vspace{-1em}
\paragraph{Text-to-image comparison.}
While the text-to-text metrics already enable good candidate selection, potential errors during candidate text generation could degrade their performance.
To alleviate this, we further perform text-to-image comparisons between the reference text and the \textit{mask cropped} visual prompt $V^m_i$.
Similar to the text-to-text comparison, we use an MLLM-generated binary decision 
$d_i^\text{t2i} = f_\text{MLLM}(T^\text{ref}, V_i^m \;|\; Q^\text{t2i}) \in \{ 0, 1 \}$, 
followed by a CLIP-generated scalar score
$s_i^\text{t2i} = f_\text{CLIP}(T^\text{ref}, V_i^m) \in [ 0, 1 ]$,
where $Q^\text{t2i}$ is the text-to-image comparison prompt as shown in the lower right corner of Figure~\ref{fig:voting}.

\vspace{-1em}
\paragraph{Grouping and selection.}
Given the computed metrics, we select the mask candidate that best matches the input expression $E$, or return the reference text $T^\text{ref}$ if none is found.
Algorithm~\ref{alg:selection} summarizes this process.

\rzz{As MLLM decisions are prioritized over CLIP sores,} we begin by checking whether any masks receive positive responses for both text-to-text and text-to-image MLLM decisions. 
In practice, we notice that the correct candidate is often the union of all the candidate masks that satisfy this condition, especially in cases where a single semantic entity spans multiple segments (e.g., all legs of a sofa). 
Therefore, we also include the union of these masks as another viable candidate.
We then return the mask candidate with the highest combined CLIP score (sum of $s_\text{t2t}$ and $s_\text{t2i}$).
If no such masks exist, we then repeat this process, using only the text-to-text MLLM decisions as the filter, and then using only the text-to-image MLLM decisions as the filter.

We \rzz{also} prioritize text-to-text over text-to-image decisions, as empirically, we find the former more reliable.
As a final verification step (lines 17-20 in Algorithm~\ref{alg:selection}), when no candidates receive positive MLLM responses, we check if any of them has a combined CLIP score over a threshold (set to $1$ for all experiments), and return the mask with the highest score. 
This threshold helps identify cases where the target is either invisible or irrelevant to the image, in which case we return the reference text $T^\text{ref}$ explanation that describes why the target cannot be localized. 

\iccv{This algorithm enables our method to handle occlusion cases by combining parts segmentations, while also generalizing to multi-object scenarios. Additional discussions and results are available in the supplementary materials.}

\begin{algorithm}
\footnotesize
\caption{Grouping and Selection Process}
\label{alg:selection}
\begin{algorithmic}[1]
\State $conditions \gets \{(True, True), (True, False), (False, True)\}$
\For{$(t2t, t2i)$ in $conditions$}
    \If{$t2t$ and $t2i$}
        \State $C \gets \{ m_i \mid d^\text{t2t}_i = 1 \land d^\text{t2i}_i = 1 \}$
    \ElsIf{$t2t$}
        \State $C \gets \{ m_i \mid d^\text{t2t}_i = 1 \}$
    \ElsIf{$t2i$}
        \State $C \gets \{ m_i \mid d^\text{t2i}_i = 1 \}$
    \EndIf
    \If{$|C| = 1$}
        \State \Return $C[0]$
    \ElsIf{$|C| > 1$}
        \State $m_\text{cmb} \gets \text{CombineMasks}(C)$
        \State Compute $s^\text{t2t}_\text{cmb}$, $s^\text{t2i}_\text{cmb}$
        \State \Return $\argmax_{m \in \{C \cup m_\text{cmb}\}} (s_\text{t2t}^m + s_\text{t2i}^m)$
    \Else
        \State pass
    \EndIf
\EndFor
\If{$\max_{m} (s^\text{t2t}_m + s^\text{t2i}_m) < 1$}
    \State \Return $T^\text{ref}$
\Else
    \State \Return $\argmax_{m\in M} (s^\text{t2t}_m + s^\text{t2i}_m)$
\EndIf
\end{algorithmic}
\end{algorithm}

\section{Experiment}
\label{sec:experiment}


We use Pixtral 12B~\cite{agrawal2024pixtral} as the MLLM, SAM ViT-H~\cite{kirillov2023segment} for generating segmentation proposals, and CLIP-ViT-B-32 for CLIP scores.
\iccv{Our experiments were conducted on a server with 8 NVIDIA 32GB V100 GPUs for parallel inference, but the entire inference process can run effectively on just a single NVIDIA 24GB 4090 GPU. Additional inference time details are provided in the supplementary materials.}

\vspace{-1em}
\paragraph{Public datasets.} 
Following the most previous works on referring segmentation~\cite{lai2024lisa, chen2025sam4mllm}, we evaluate the performance of \ourmethod{} on four public benchmark datasets: RefCOCO, RefCOCO+~\cite{yu2016modeling}, RefCOCOg~\cite{mao2016generation, nagaraja2016modeling} and ReasonSeg~\cite{lai2024lisa}. Being a zero-shot method, we directly evaluate on the validation and test sets without any fine-tuning. 

\begin{figure}[t]
    \centering
    \includegraphics[width=\linewidth]{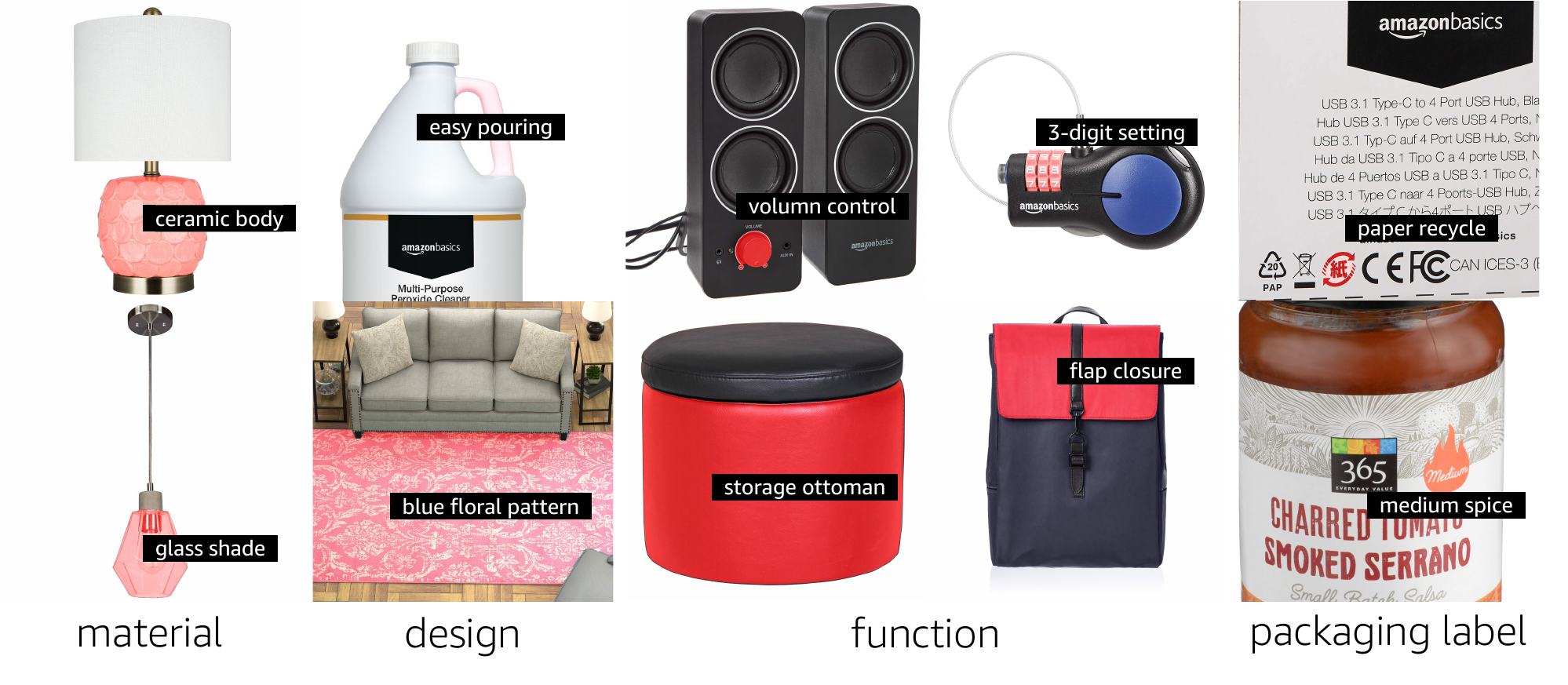}
    \vspace{-20pt}
    \caption{Examples of different expressions in \ourdataset. Best viewed with zoom-in.}
    \label{fig:abo-vis}
    \vspace{-10pt}
\end{figure}
\begin{table*}[t]
\caption{Quantitative results on standard RES benchmarks refCOCO/+/g, reported as cIoU values.}
\label{tab:quan-refcoco}
\centering
\footnotesize
\begin{tabular}{lcccccccccccc}
\toprule
\multicolumn{1}{l}{\textbf{Method}} & & \multicolumn{3}{c}{\textbf{refCOCO}} & &  \multicolumn{3}{c}{\textbf{refCOCO+}} & & \multicolumn{3}{c}{\textbf{refCOCOg}} \\ \cmidrule{3-5} \cmidrule{7-9} \cmidrule{11-13}
\multicolumn{1}{c}{} & & val & testA & testB & & val & testA & testB & & val(U) & val(G) & test(U) \\ \midrule
\textit{fully-supervised on the training set} & \\
VLT~\cite{ding2021vision} & & 67.5 & 70.5 & 65.2 &  & 56.3 & 61.0 & 50.1 &  & 55.0 & - & 57.7\\
CRIS~\cite{wang2022cris} & & 70.5 & 73.2 & 66.1 &  & 62.3 & 68.1 & 53.7 &  & 59.9 & - & 60.4 \\
LAVT~\cite{yang2022lavt} & & 72.7 & 75.8 & 68.8 & & 62.1 & 68.4 & 55.1 & & 61.2 & - & 62.1 \\
GRES~\cite{liu2023gres} &  & 73.8 & 76.5 & 70.2 &  & 66.0 & 71.0 & 57.7 &  & 65.0 & - & 66.0 \\
\midrule
\emph{pre-trained on the same task} \\
UniRES~\cite{wang2024unveiling} & & 71.2 & 74.8 & 66.0 & & 59.9 & 66.7 & 51.4 & & 62.3 & - & 63.2\\
LISA-7B~\cite{lai2024lisa} & & 74.9 & 79.1 & 72.3 &  & 65.1 & 70.8 & 58.1 &  & 67.9 & - & 70.6 \\
GSVA~\cite{xia2024gsva} & & 77.2 & 78.9 & 73.5 & & 65.9 & 69.6 & 59.8 & & 72.7 & - & 73.3 \\
GLaMM~\cite{rasheed2024glamm} & & 79.5 & \textbf{83.2} & \textbf{76.9} & & 72.6 & 78.7 & 64.6 & & 74.2 & - & 74.9 \\ 
\rowcolor{red!10!white}SAM4MLLM~\cite{chen2025sam4mllm}& & \textbf{79.8} & 82.7 & 74.7 & & \textbf{74.6} & \textbf{80.0} & \textbf{67.2} & & \textbf{75.5} & - & \textbf{76.4} \\
\midrule
\textit{training-free zero-shot} &  \\
GLCLIP~\cite{yu2023zero} & & 26.2 & 24.9 & 26.6 & & 27.8 & 25.6 & 27.8 & & 33.5 & 33.6 & 33.7 \\
CaR~\cite{sun2024clip} & & 33.6 & 35.4 & 30.5 & & 34.2 & 36.0 & 31.0 & & 36.7 & 36.6 & 36.6\\

\rowcolor{blue!10!white}\ourmethod & & \textbf{68.5} & \textbf{72.2} & \textbf{70.3} & & \textbf{60.7} & \textbf{65.6} & \textbf{52.2} & & \textbf{60.1} & \textbf{60.5} & \textbf{60.9} \\
\bottomrule
\end{tabular}
\vspace{-1em}
\end{table*}

\vspace{-1em}
\paragraph{\ourdataset~benchmark.} 
To further evaluate the capability of \ourmethod{} in handling implicit expressions (e.g., part-level materials, features, and functionalities), we establish the \ourdataset{} benchmark for complex reasoning segmentation tasks. We build upon the ABO dataset, which contains product listings with rich metadata, images, and 3D models from Amazon.com. Our benchmark comprises 2,482 high-resolution catalog images spanning 565 product types, with 2,989 referring expressions targeting part-level regions that describe specific materials, features, functionalities, or packaging elements. Fig.~\ref{fig:abo-vis} shows representative examples, with detailed refer extraction procedures and data annotation provided in the supplementary.


\begin{table}[t]
\caption{Quantitative results on ReasonSeg.}
\vspace{-0.5em}
\label{tab:quan-reasonseg}
\centering
\footnotesize
\begin{tabular}{llll}
\toprule
\multirow{2}{*}{Method} & & \multicolumn{2}{c}{val} \\\cmidrule{3-4}
&  & \multicolumn{1}{c}{gIoU}  & \multicolumn{1}{c}{cIoU}\\
\midrule
GLaMM \cite{rasheed2024glamm} & & 47.4 & 47.2\\
LISA-7B-LLaVA1.5 \cite{lai2024lisa} & & 53.6 & 52.3  \\
LISA-13B-LLaVA1.5 \cite{lai2024lisa} & & 57.7 & 60.3 \\
SAM4MLLM~\cite{chen2025sam4mllm} & & 58.4 & 60.4 \\
\rowcolor{blue!10!white}\ourmethod & & 74.6 & 72.5\\
\bottomrule
\end{tabular}
\end{table}
\begin{table}[t]
\caption{Quantitative results on \ourdataset.}
\vspace{-0.5em}
\label{tab:quan-abo}
\centering
\footnotesize
\begin{tabular}{llll}
\toprule
\multirow{2}{*}{Method} & & \multicolumn{2}{c}{test} \\\cmidrule{3-4}
&  & \multicolumn{1}{c}{gIoU}  & \multicolumn{1}{c}{cIoU}\\
\midrule
LISA-13B-LLaVA1.5~\cite{lai2024lisa} & & 43.3 & 34.0 \\
GLaMM~\cite{rasheed2024glamm} & & 46.2 & 38.7\\
\rowcolor{blue!10!white}\ourmethod & & 78.2 & 72.4\\
\bottomrule
\vspace{-3em}
\end{tabular}
\end{table}

\vspace{-1em}
\paragraph{Evaluation metrics.}
We evaluate our method using two standard metrics following prior works~\cite{lai2024lisa, rasheed2024glamm}: generalized IoU (gIoU) and cumulative IOU (cIoU). gIoU computes the average of per-image Intersection-over-Union scores, while cIoU measures the ratio of cumulative intersection to cumulative union across all images. We report gIOU for RefCOCO, RefCOCO+, and RefCOCOg, and both metrics for ReasonSeg and \ourdataset.


\subsection{Evaluation on Vanilla RES}
We evaluate \ourmethod{} on standard referring segmentation benchmarks, as shown in Table~\ref{tab:quan-refcoco}. Our method significantly outperforms existing zero-shot approaches, more than doubling the performance of GLCLIP (68.5\% vs 26.2\% on refCOCO val set) and achieving comparable results with early supervised methods like VLT. \iccv{Despite UniRES~\cite{wang2024unveiling} being described as a zero-shot method, it was pre-trained on their proposed MRES-32M dataset, which remains unavailable to the public. Furthermore, due to UniRES being closed source, our comparisons are limited to the accuracy figures reported in their paper.} The performance gap compared to recent supervised methods can be attributed to our segmentation strategy with smaller mask proposals, which faces challenges when handling large complete objects that are common in these datasets. Qualitative results are provided in the supplementary. \iccv{Furthermore, we evaluate \ourmethod{} with competing methods on more general part-level and multi-object referring segmentation tasks as detailed in our supplementary.}

\subsection{Evaluation on Reasoning Segmentation}
\begin{figure*}[t!]
    \centering
    \begin{minipage}{0.48\textwidth}
        \centering
        \includegraphics[width=\linewidth]{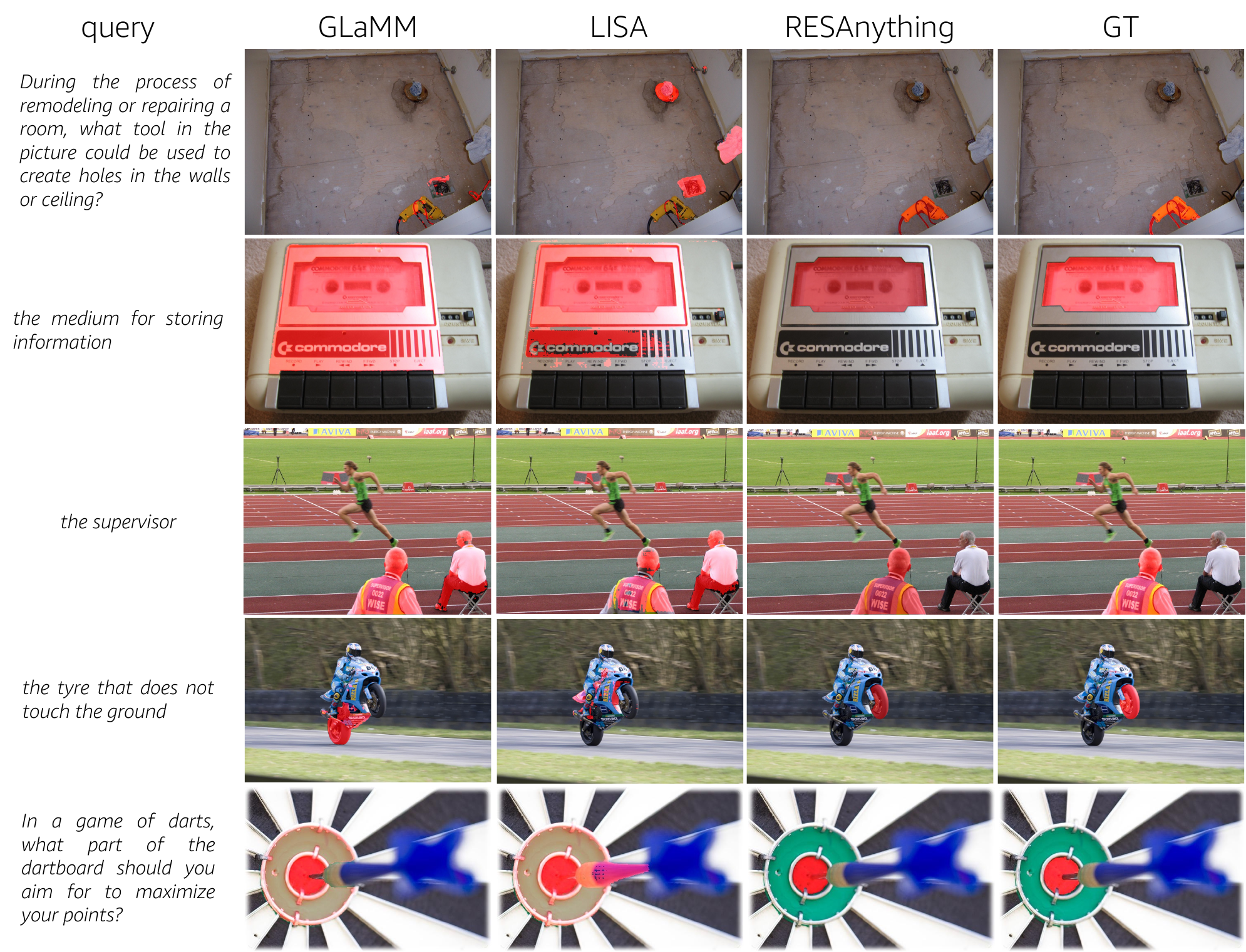}
        \caption{Qualitative comparisons on ReasonSeg. Our method demonstrates superior performance in both object localization accuracy (rows 1, 3, 4) and segmentation precision (rows 2, 5). }
        \label{fig:qual-ReasonSeg}
        \vspace{-1em}
    \end{minipage}
    \hfill
    \begin{minipage}{0.48\textwidth}
        \centering
        \includegraphics[width=\linewidth]{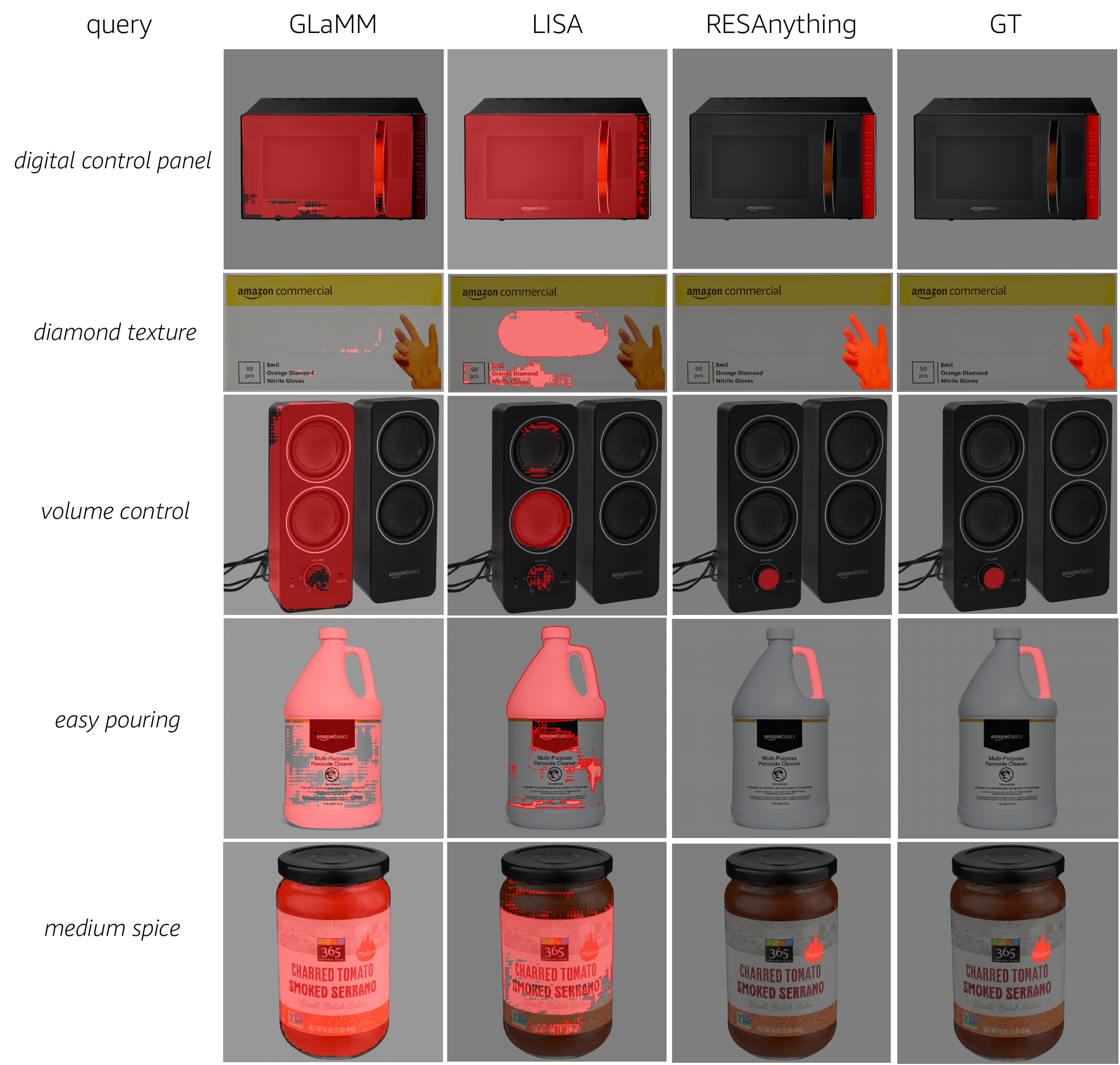}
        \caption{Qualitative comparisons on \ourdataset. \ourmethod{} demonstrates superior generalization ability across diverse queries, producing more fine-grained segmentation.}
        \label{fig:qual-abo}
        \vspace{-1em}
    \end{minipage}
\end{figure*}

We evaluate \ourmethod{} on the ReasonSeg benchmark (Table~\ref{tab:quan-reasonseg}), where our method achieves state-of-the-art performance of 74.6\% gIoU and 72.5\% cIoU, surpassing LISA-13B by 17\% and SAM4MLLM by 16\%. Notably, while LISA variants require fine-tuning on reasoning tasks and GLaMM \& SAM4MLLM rely on extensive training data, \ourmethod{} achieves this superior performance without any task-specific training, demonstrating the effectiveness of leveraging MLLMs for deep reasoning. Qualitative comparisons are shown in Fig~\ref{fig:qual-ReasonSeg}.

\ourdataset{} contains more challenging referring expressions targeting materials, features, functionalities or package elements. On this benchmark, \ourmethod{} achieves 78.2\% gIoU and 72.4\% cIoU, significantly outperforming GLaMM by over 30\% in both metrics, demonstrating our method's strong capability in handling complex reasoning queries. Qualitative comparisons are shown in Fig~\ref{fig:qual-abo}.

\subsection{Ablation Study}
\paragraph{Visual prompts.}
As shown in Fig~\ref{fig:visual-prompt}, we explore different types of visual prompts for generating candidate texts $T^{can}$ and performing text-to-image comparison. Table~\ref{tab:ablation-vp} compares their performance on RefCOCO test A set. The combination of mask-cropped and bounding box prompts achieves the best performance (72.2\% gIoU), while using mask alone yields the lowest (47.2\% gIoU) as it obscures contextual relationships. This demonstrates the importance of preserving spatial context through bounding box while maintaining region-specific details through mask cropping. Additional analysis is provided in the supplement.

\vspace{-1em}
\paragraph{MLLM backbone.}
To analyze the impact of varying the MLLM backbone, we compare the performance of different MLLMs on ReasonSeg. Table~\ref{tab:ablation-llm} summarizes the results. While Pixtral-12B is our default choice, both Qwen2-VL and Claude 3.5 Sonnet achieve comparable or slightly better performance (74.2-76.2\% gIoU), demonstrating our method's robustness across different MLLMs. See supplementary materials for extended analysis.

\begin{table}[t]
\vspace{-1em}
\caption{Ablation study on different visual prompts.}
\vspace{-0.5em}
\label{tab:ablation-vp}
\centering
\resizebox{0.95\columnwidth}{!}{
\footnotesize
\begin{tabular}{cccccccc}
\toprule
\multirow{2}{*}{Dataset} & \multicolumn{5}{c}{Visual Prompts} & \multirow{2}{*}{gIoU} & \multirow{2}{*}{cIoU}\\
& image & mask & bbox & contour & blur & \\
\midrule
& & \checkmark & & & & 47.2 & 42.3\\
& \checkmark & \checkmark & & & & 56.2 & 53.3\\
& \checkmark & & & \checkmark & & 48.4 & 44.2\\
RefCOCO& & & & & \checkmark & 43.5 & 39.2 \\
test A & & \checkmark & & & \checkmark & 67.4 & 64.1\\
 & & \checkmark & \checkmark & & & 72.2 & 69.5\\
& & \checkmark & & \checkmark & & 68.5 & 64.4\\
& & & \checkmark & \checkmark & & 50.4 & 46.6\\
\bottomrule
\vspace{-1em}
\end{tabular}
}
\end{table}
\begin{table}[t]
\vspace{-1em}
\caption{Ablation study on MLLM backbone.}
\vspace{-0.5em}
\label{tab:ablation-llm}
\centering
\footnotesize
\begin{tabular}{lccc}
\toprule
LLM & & gIoU & cIoU\\
\midrule
Pixtral 12B\cite{agrawal2024pixtral}& & 74.6 & 72.5\\
Claude3.5 Sonnet\cite{claude-sonnet} & & 76.2 & 73.4\\
Qwen 2-VL\cite{bai2023qwen} & & 74.2 & 72.1 \\
\bottomrule
\vspace{-0.5em}
\end{tabular}
\vspace{-2em}
\end{table}

%

\section{Conclusion, limitation, and future work}
\label{sec:conclusion}

We present~\ourmethod, a zero-shot approach to advance open-vocabulary RES by supporting language expressions referring to \rz{highly general} concepts. Our method comprises two key components: a novel attribute prompting technique to extract detailed attributes as text descriptions by synergizing SAM and MLLM for CoT analysis, and a multi-metric mask selection module based on CLIP and MLLM to select the optimal mask from SAM proposals.

Our method demonstrates superior performance over prior zero-shot methods on standard RES benchmarks (RefCOCO/+/g). More importantly, our training-free approach substantially outperforms existing fine-tuned MLLM methods on both ReasonSeg~\cite{lai2024lisa} for reasoning segmentation and our newly augmented ABO dataset, underscoring its \rz{comprehensive} reasoning capabilities. 
\rz{While~\ourmethod~also performs well on object-level RES, attribute prompting excels especially at part-level reasoning since the attributes considered (color, shape, and location) tend to exhibit more consistency over parts, than objects, that share similar functions, styles, material, etc. It would be interesting to explore other attributes for CoT or automate the prompts.}

\iccv{Our method has substantial room for inference efficiency optimization in future work, particularly through RoI filtering and size-based mask proposal pruning to reduce candidate text generation overhead.}
\ourmethod~also inevitably inherits limitations common to foundation model-based approaches. Notably, SAM occasionally fails to produce the best mask candidates, potentially 
degrading RES accuracy, \iccv{as shown in the supplementary materials.}
In addition, the effectiveness of \ourmethod~depends on the specific MLLMs employed. Future work could focus on improving the mask proposal generation process and exploring the integration of more advanced LLMs/MLLMs.

\section*{Acknowledgements}
We thank Yiming Qian, Kai Wang, Fenggen Yu for their invaluable contributions in the early stage of this project.

\clearpage
\setcounter{page}{1}
\maketitlesupplementary
The supplementary document provides (1) detailed analysis of limitation of current methods, including both MLLM and CLIP in our task, in Section~\ref{sec:supp-limitation}; (2) comprehensive details of language and visual prompts used in \ourmethod~in Section~\ref{sec:prompts}; (3) additional information about the construction of \ourdataset~in Section~\ref{sec:dataset}; (4) extended quantitative \iccv{results on part level and multi-object GRES task} and qualitative \iccv{results, including failure cases} in Section~\ref{sec:quan} and \ref{sec:qual}, respectively. 

\section{Limitation of Current Methods}
\label{sec:supp-limitation}
Our method leverages Chain-of-Thought (CoT) attribute prompting for detailed descriptions and combines MLLMs and CLIP as mask selector to select optimal segmentation proposals. While this dual-model approach achieves strong performance, it arises from the inherent limitations of both components. In this section, we analyze the constraints of current MLLMs and CLIP that motivate our design choices in attribute prompting and the hybrid evaluation strategy.
\subsection{Limitation of MLLM}
\begin{figure*}
    \centering
    \includegraphics[width=\linewidth]{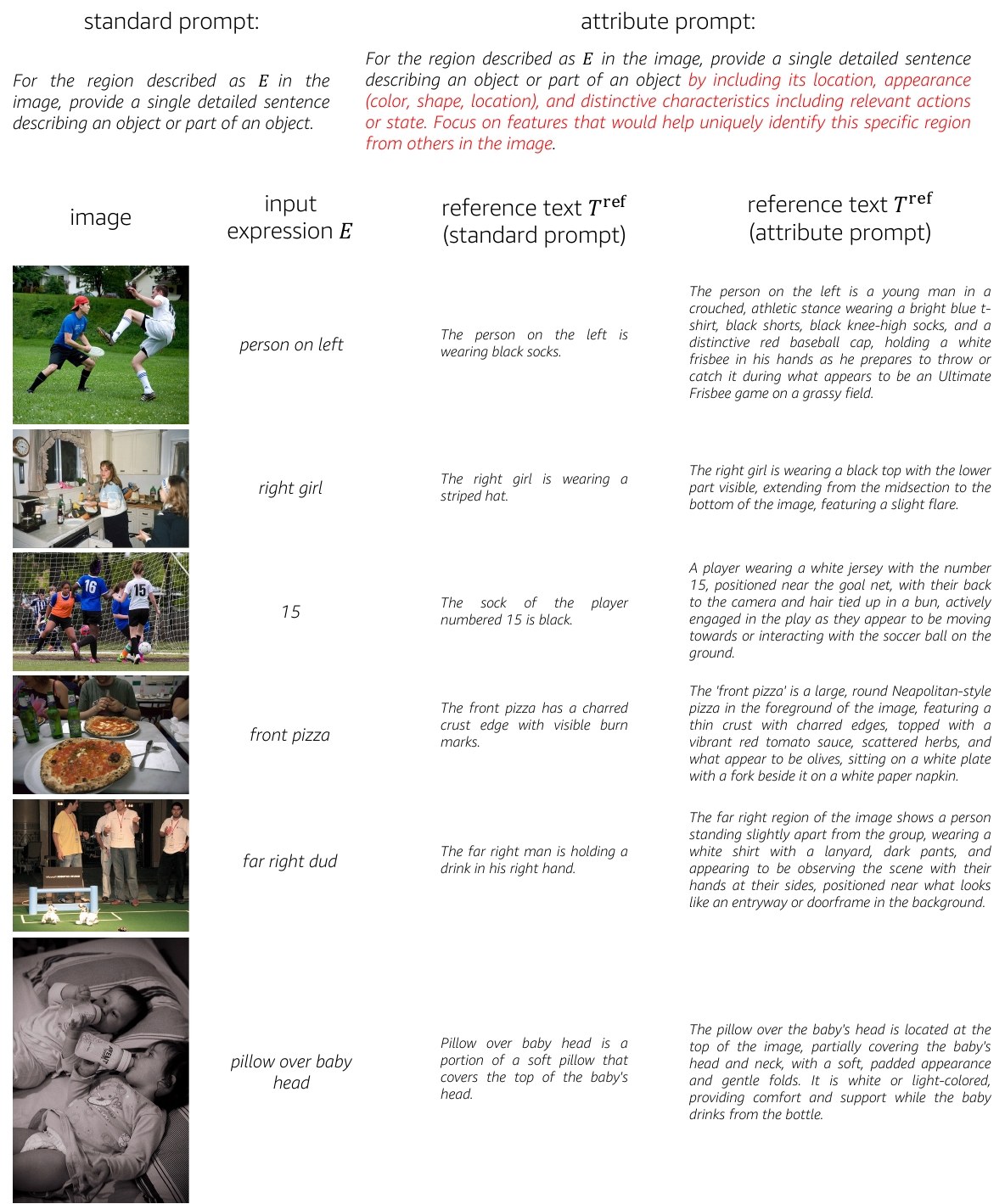}
    \caption{Comparison of Text Generation With and Without Attribute Prompting: Our analysis demonstrates that when attribute prompting is not used, MLLM fails to accurately identify and reason about input expression attributes. The contrast between standard prompting and attribute-specific prompting highlights this significant limitation in attribute recognition.}
    \label{fig:mllm-limitation}
\end{figure*}

\begin{figure*}
    \centering
    \includegraphics[width=0.87\linewidth]{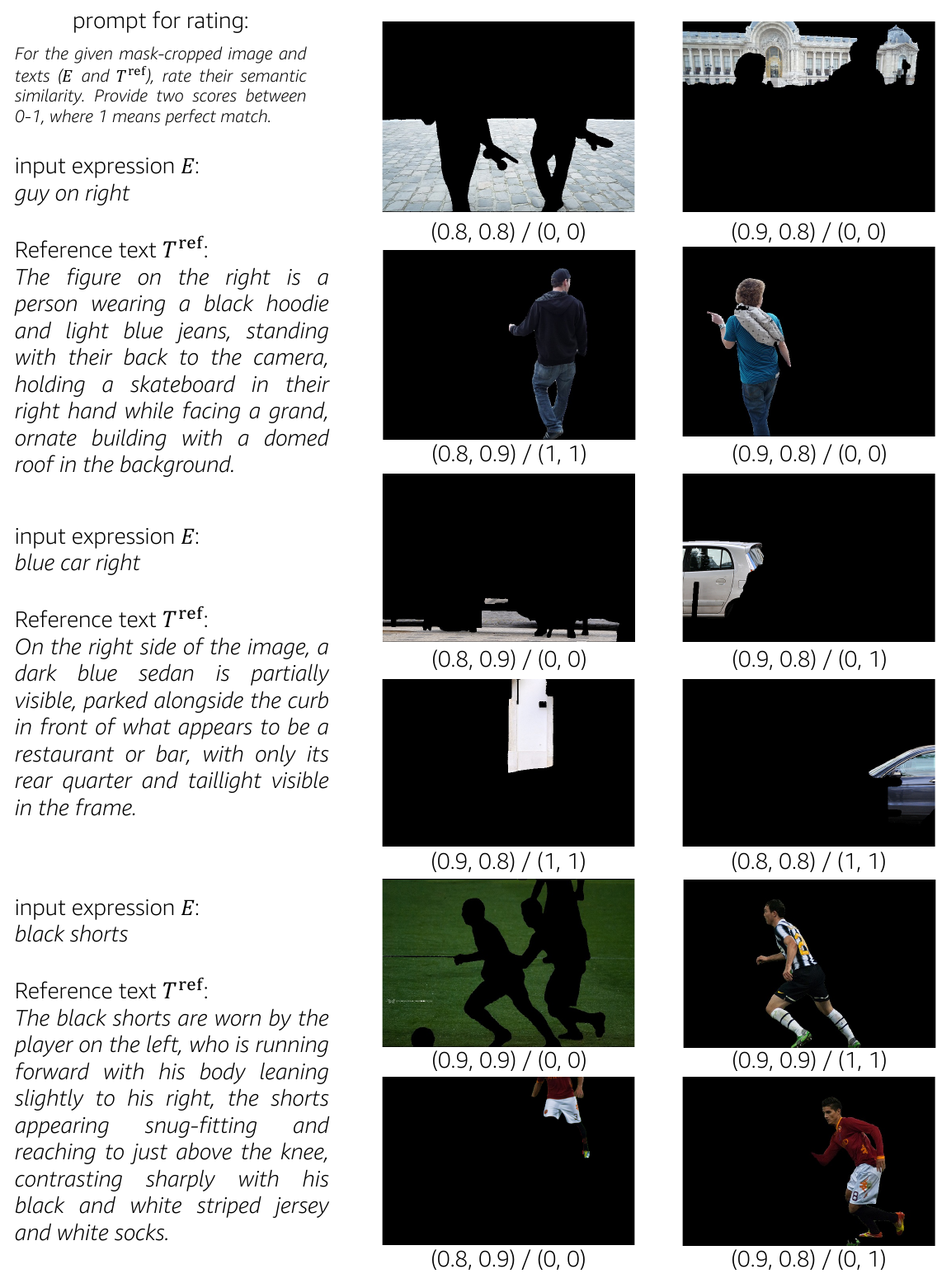}
    \caption{Analysis of MLLM's Rating and Binary Response Performance: For each mask-cropped region, we compare two types of outputs: numerical scores (score 1, score 2) and binary responses ($d^\text{t2t}$, $d^\text{t2i}$) (0='no', 1='yes'). The results reveal that MLLM struggles to generate meaningful similarity scores when comparing the input expression $E$ and reference text $T^\text{ref}$. The assigned scores (typically around 0.8-0.9) appear arbitrary rather than reflecting accurate contextual similarities. In contrast, the model's binary yes/no responses prove more reliable for assessment purposes.}
    \label{fig:mllm-rating}
\end{figure*}

\paragraph{Attribute Prompt.}
While MLLMs exhibit strong reasoning capabilities, they often fail to perform systematic CoT reasoning without explicit prompting guidance. As shown in Fig~\ref{fig:mllm-limitation}, when asked to describe the details of input expression $E$ without specific attribute requirements, MLLMs typically generate oversimplified descriptions that fail to capture the target's essential characteristics and details effectively. Therefore, providing MLLMs with explicit attribute requirements is essential to guide their reasoning process effectively. \ourmethod~leverages this insight to generate more comprehensive and accurate descriptions, ensuring that all necessary details of the target expression are properly captured.
\paragraph{Binary Response.}
As mentioned in our main paper, a naive approach for applying MLLMs to solve our task would involve prompting the MLLMs to output a score for each segmentation proposal $m_i$, denoting its similarity with the input expression $E$. However, MLLMs are primarily designed to understand and generate text rather than compute precise numerical similarities. While they excel at comparing and reasoning about content qualitatively, they struggle to produce reliable numerical similarity scores. Our experiments reveal that MLLM-generated similarity scores exhibit high variance and poor correlation with actual contextual similarity, as the model essentially samples from its probability distribution rather than performing true similarity computation. Therefore, we reformulate similarity assessment as binary classification queries, returning yes or no in our selection algorithm, which better aligns with MLLMs' natural language understanding capabilities. As shown in Fig~\ref{fig:mllm-rating}, our experiments reveal that MLLMs tend to generate similarity scores that appear arbitrary or biased by their training distribution, rather than computing true similarities between the given elements, and their binary responses prove to be more reliable indicators.

\subsection{Limitation of CLIP}
\begin{figure*}
    \centering
    \includegraphics[width=0.9\linewidth]{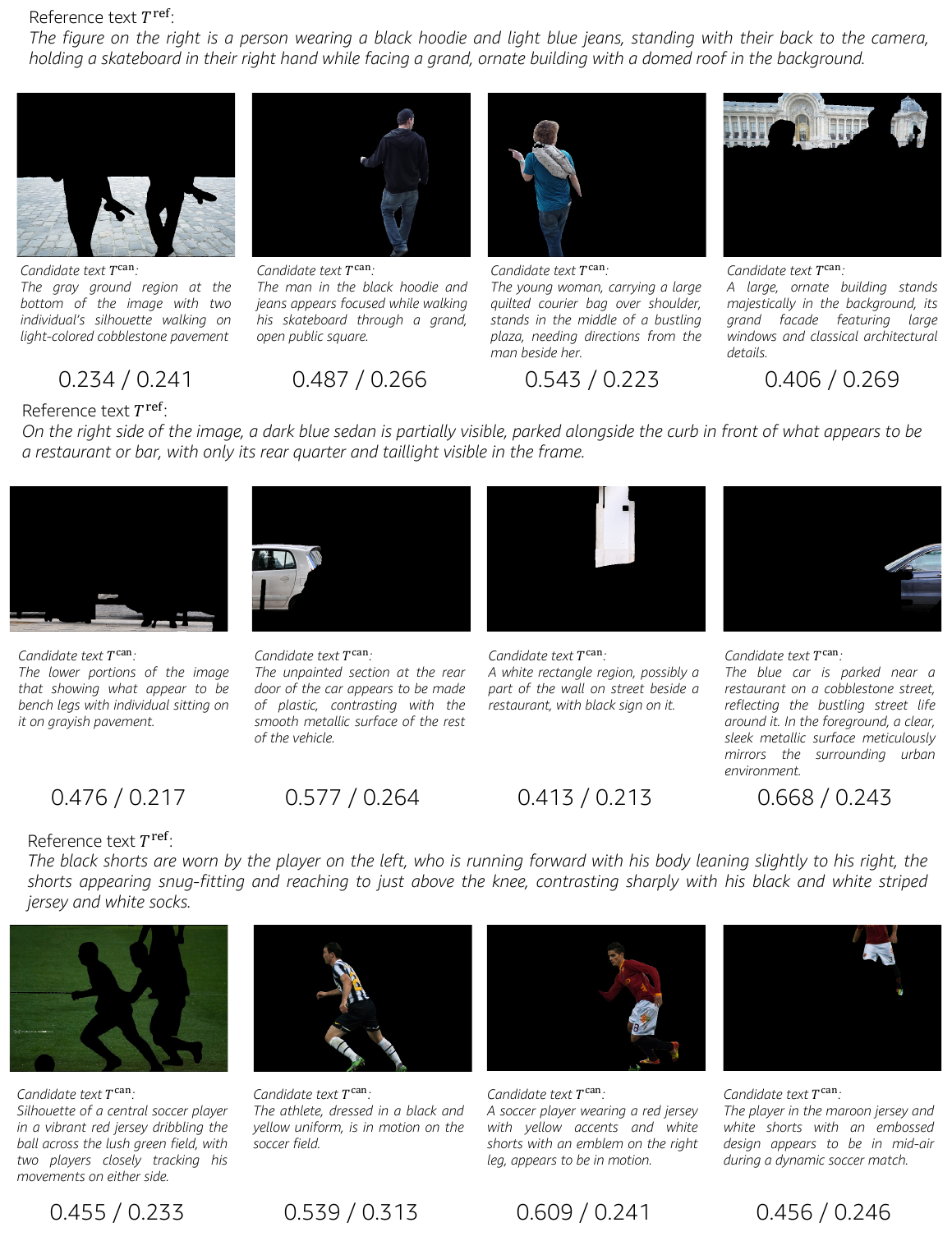}
    \caption{Analysis of CLIP's Similarity Evaluation: For each mask-cropped region, we compare text-to-text ($s^\text{t2t}$) and text-to-image ($s^\text{t2i}$) CLIP scores. Text-to-text scores prove more reliable, while text-to-image scores consistently remain below 0.3, showing limited discriminative power. However, relying solely on text-to-text scores can be misleading, as demonstrated in the last row where a description containing "white shorts" receives a higher score despite incorrectly matching the reference image showing "black shorts". This highlights the limitation of text-to-text evaluation in capturing crucial contextual details.}
    \label{fig:clip-score}
\end{figure*}

The limitations of CLIP in analyzing contextual similarities become evident when dealing with complex descriptions and image content. As shown in Fig~\ref{fig:clip-score}, while CLIP's text-to-text similarity scores reveal meaningful comparison, they often fail to capture crucial contextual details like color attributes. Additionally, CLIP's text-to-image similarity scores show limited discriminative power, consistently remaining below 0.3. These limitations underscore our decision to adopt MLLMs as our primary mask selector, as they demonstrate superior capability in understanding and comparing detailed contextual content. 

\subsection{Ablation Study}
We further evaluate the effectiveness of adopting both MLLM and CLIP as mask selectors in \ourmethod. Table~\ref{tab:ablation-selection} compares the performance of \ourmethod~on ReasonSeg test set with different mask selectors configurations. Using CLIP as the sole mask selector results in poor performance due to its previously mentioned limitations in understanding complex relationships and abstract concepts. While MLLM demonstrates superior reasoning and contextual similarity capabilities compared to CLIP, using MLLM alone can lead to incomplete region selection, particularly for expressions targeting multiple parts (e.g., sofa legs or armrests). These results validate our design choice of incorporating both MLLM and CLIP as mask selectors to ensure robust region selection.

\begin{table}[h]
\caption{Ablation study on different mask selectors. }
\vspace{-0.5em}
\label{tab:ablation-selection}
\centering
\footnotesize
\begin{tabular}{llcc}
\toprule
\multirow{2}{*}{Method} & & \multicolumn{2}{c}{test} \\\cmidrule{3-4}
&  & \multicolumn{1}{c}{gIoU}  & \multicolumn{1}{c}{cIoU}\\
\midrule
CLIP only & & 42.5 & 38.4\\
LLM only & & 70.5 & 64.6\\
both & & 74.6 & 72.5 \\
\bottomrule
\end{tabular}
\end{table}

\section{Prompts}
\label{sec:prompts}
\subsection{Language Prompts in Attribute Prompting}
As mentioned in the main paper, we use reference text prompt $Q^\text{ref}$ to generated reference text $T^\text{ref}$ for each refer based on the input expression $E$. Given the input image $I$ and referring expression $E$, we prompt the MLLM using following $Q^\text{ref}$ to obtain reference text $T^\text{ref}$: 
\\ \\
\noindent\texttt{For the region described as \{$E$\} in the image, provide a single detailed sentence describing an object or part of a object by including its location, appearance (color, shape, location), and distinctive characteristics including relevant actions, state, or function. Focus on features that would help uniquely identify this specific region from others in the image. Be as succinct as possible and in English only.}
\\ \\
Similarly, given the mask cropped $V^m_i$ and bounding box image $V^b_i$ as visual prompts of a segmentation proposal $m_i$ we prompt the MLLM using following $Q^\text{can}$ to obtain candidate text $T^\text{can}$:
\\ \\
\noindent\texttt{You are presented with two complementary views of the same region: 1) A cropped masked view showing detailed visual properties; 2) A full view with a bounding box showing location and context. Generate a single detailed sentence following these guidelines: }

\noindent\texttt{FOR COMPLETE OBJECTS: \\
- Combine visual details and spatial context naturally; \\
- Visual properties (color, shape, texture, size); \\
- Location in the scene; \\
- Relationships with surroundings;\\
- State or action if relevant; }

\noindent\texttt{FOR PARTIAL REGIONS: \\
Describe the part while providing clear context: \\
- Part identification and its visual properties; \\
- Its position within the larger object/scene; \\
- Relevant contextual details; \\ 
Important Rules: Start directly with the subject: 'A [description]...' or 'The [description]...'; \\
Describe only what is visible in the non-black regions for visual properties and the image with green bounding box is for location and relation analysis; \\
Never mention masks, boxes, or annotations; \\
Use confident language for clear identifications; \\
Use tentative language when inferring; \\
Create natural, flowing descriptions that combine all information seamlessly; \\
Focus on creating cohesive descriptions that feel natural and informative without drawing attention to the source of the information.}
\\\\
We adjust the $Q^\text{can}$ based on different visual prompts for ablation study, e.g. mask cropped $V^m_i$ only: \texttt{You are presented with a cropped masked view showing detailed visual properties; ...}

\begin{figure*}
    \centering
    \includegraphics[width=\linewidth]{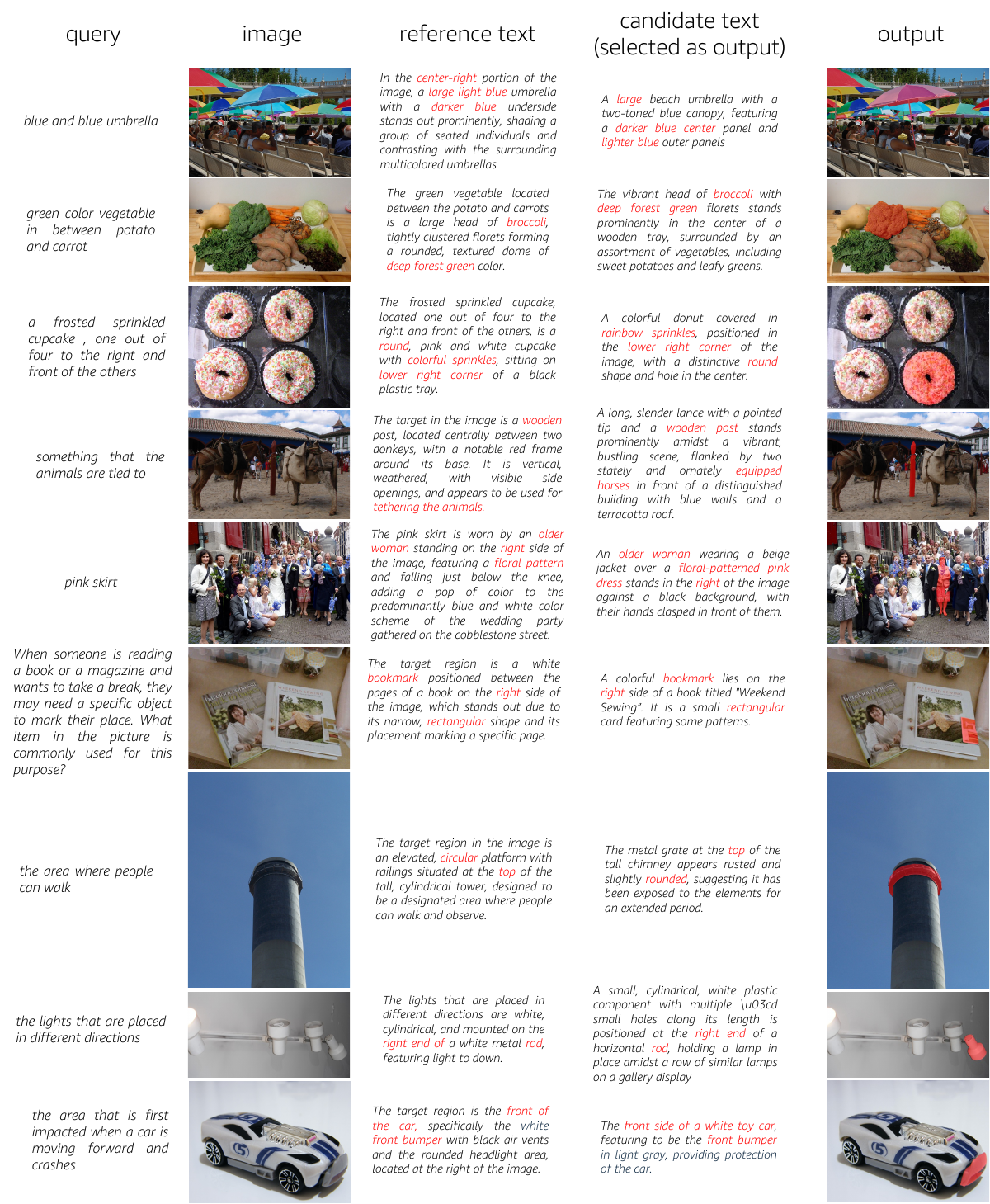}
    \caption{Examples of query, reference, and candidate text. For each input expression query (column 1), \ourmethod~generates detailed reference text describing the input expression's attributes (column 3). Our grouping and selection algorithm identifies the most relevant segmentation from candidates. Columns 4 and 5 show \ourmethod's output segmentation and its corresponding candidate text. Key words of attributes in both texts are highlighted in red color.}
    \label{fig:ref-and-can-text}
\end{figure*}

Fig~\ref{fig:ref-and-can-text} shows examples of query (input expression) and generated reference \& candidate text. 

\subsection{Language Prompts in Grouping and Selection}
We employ MLLM as one of the mask selectors in our grouping and selection algorithm. Certainly, for text-to-text decision $d^\text{t2t}$, we use following $Q^\text{t2t}$:
\\\\
\noindent\texttt{You are evaluating if the following candidate text describes the input expression region:  $E$. Reference information provided for context if the input expression text is not clear: $T^\text{ref}$. Here is the candidate text to evaluate: $T^\text{can}$. \\
Evaluate if the candidate text refer to the target by checking: \\
- Spatial location match; \\
- Visual characteristics match (color, shape, size); \\
- Object/subject identity match; \\
- State/action consistency (if applicable). \\
Return 'yes' or 'no' ONLY: 'yes' if most aspects substantially match; 'no' if some significant aspect differs.}
\\\\
For text-to-image decision $d^\text{t2i}$, we use following $Q^\text{t2i}$:
\\\\
\noindent\texttt{You are evaluating if the following reference text describes the non-black region of the cropped mask image: $T^\text{ref}$. The target is $E$ for context if the reference text is inaccurate. You have two images for context: 1) A cropped mask image showing a region in non-black color; 2) An image with a green bounding box surrounding the region showing the full scene and spatial relationships. Evaluate if the reference text describes the non-black region of the cropped mask image by checking: \\
- Spatial location match (the location is relative location, not absolute location); \\
- Visual characteristics match (color, shape, size) \\
- Object/subject identity match (the masked image could be only a part of the target); \\
- State/action consistency (if applicable). \\
Return 'yes' or 'no' ONLY: 'yes' if most aspects substantially match; 'no' if some significant aspect differs.}
\subsection{Visual Prompts Selection}
\begin{figure*}
    \centering
    \includegraphics[width=\linewidth]{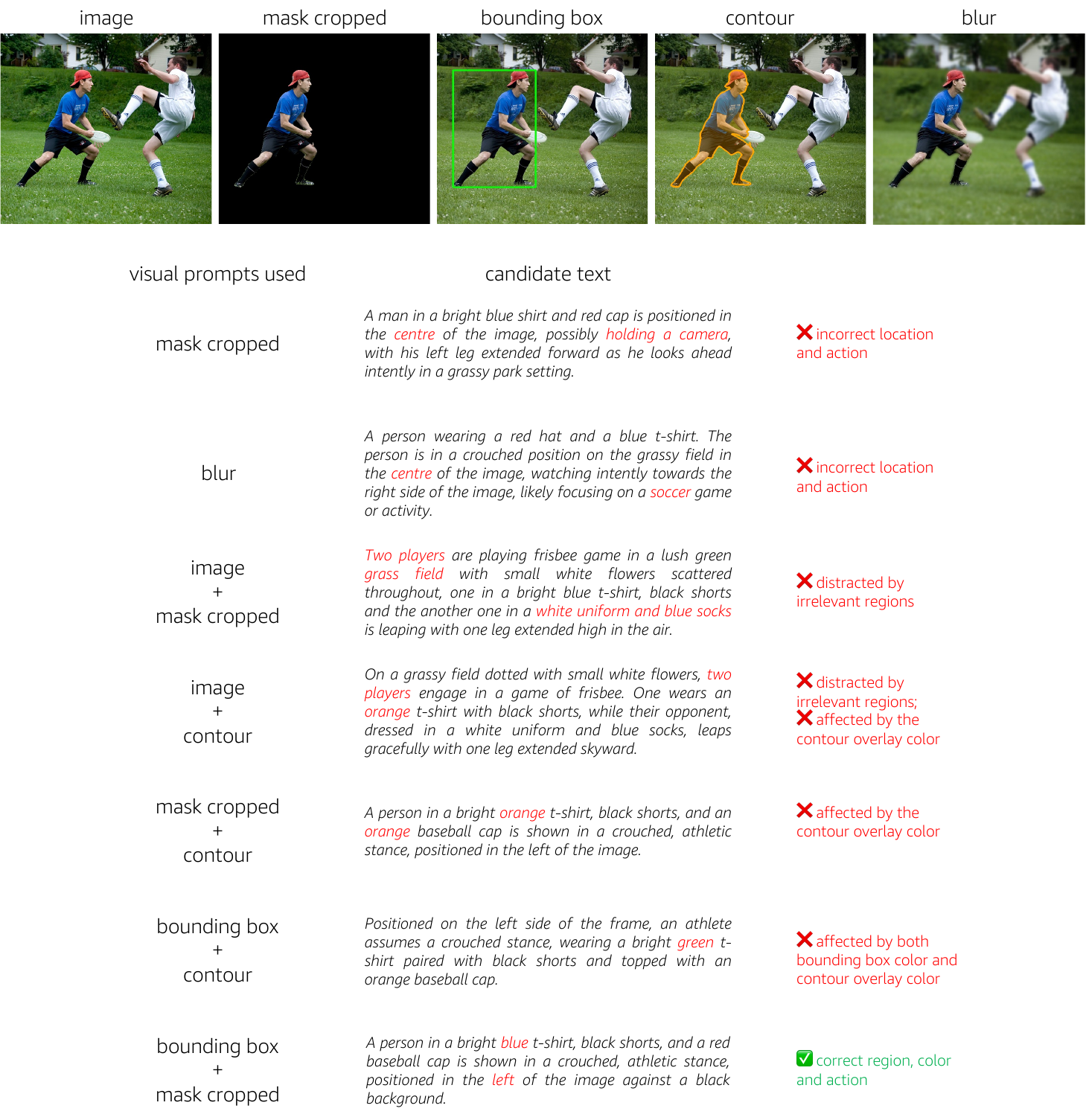}
    \caption{Comparison of different visual prompt combinations for attribute description generation. Top row shows the four basic visual prompts: original image, mask-cropped region, bounding box overlay, contour overlay and blur background. Bottom rows demonstrate how different combinations affect MLLM's generated descriptions. Using mask-cropped or blur alone leads to incorrect location and action inference, while combining with original image causes distraction from irrelevant regions. Contour-based approaches (with either mask-cropped or bounding box) suffer from color overlay interference. Our chosen combination of bounding box and mask-cropped achieves the most accurate descriptions by leveraging spatial context while avoiding color interference.}
    \label{fig:visual-prompts-compare}
\end{figure*}

\begin{figure*}
    \centering
    \includegraphics[width=0.95\linewidth]{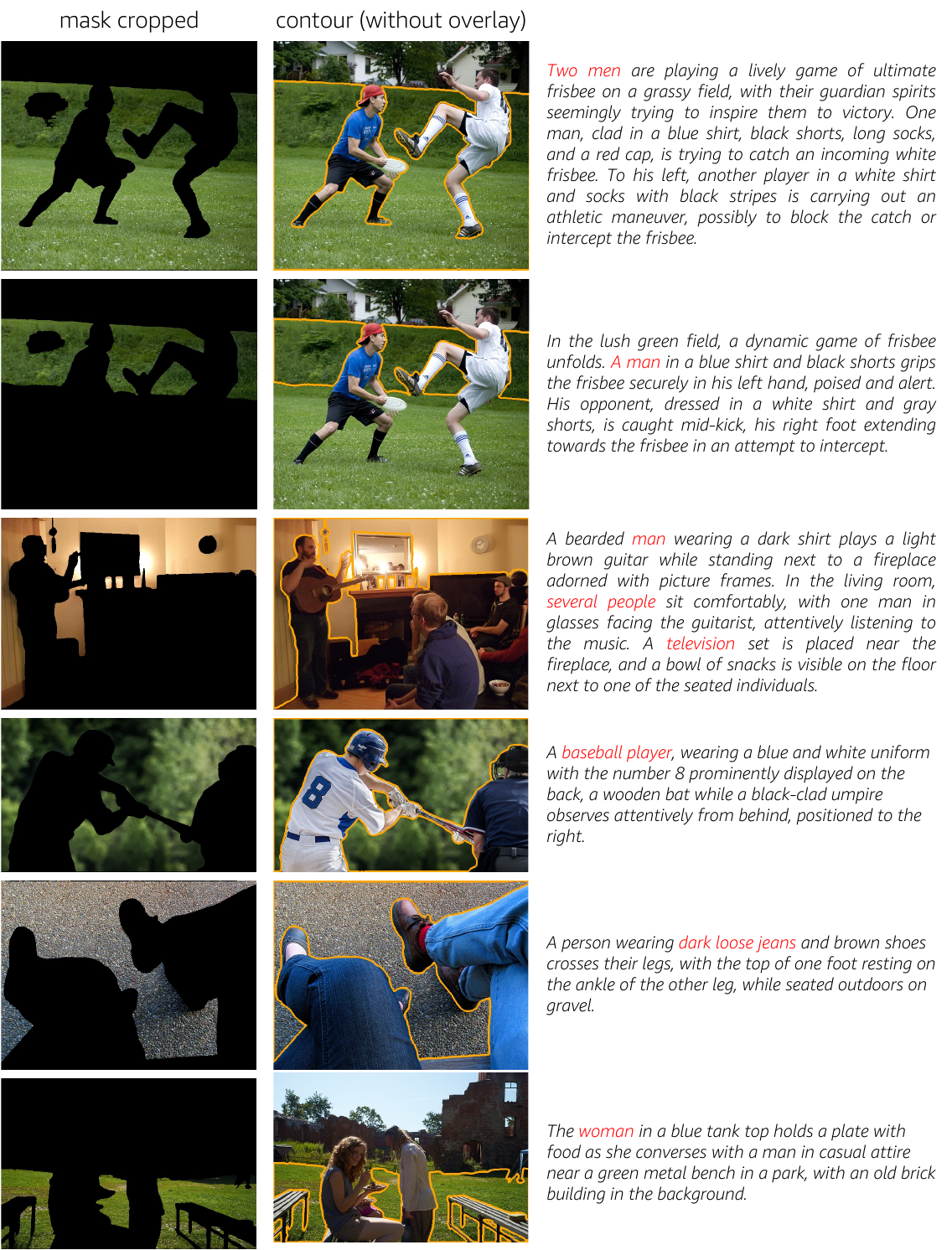}
    \caption{When dealing with non-convex shapes, analyzing only the contour without considering the overlaid mask region can lead to ambiguous visual interpretations. This ambiguity often results in generated text descriptions that contain misleading information, where incorrectly identified objects are highlighted in red.}
    \label{fig:overlay-only}
\end{figure*}

We explore five visual prompts $V_i$ in our method: (1) original image, (2) mask-cropped image, (3) bounding box overlaid on image, (4) mask contour overlaid on image and (5) blur background overlaid on image. We choose the combination of mask-cropped image and bounding box overlaid on image as the best visual prompts $V_i$ to obtain candidate text $T^\text{can}$. Apart from quantitative results presented in the ablation study, we further analyze the effectiveness and limitation of different individual/combinations of these visual prompts, as shown in Fig~\ref{fig:visual-prompts-compare}:
\begin{itemize}
    \item mask cropped only: with mask cropped as the only visual prompt, MLLM is usually failed to infer the action/relation of the region. Example in Fig~\ref{fig:visual-prompts-compare} shows that from mask cropped image, MLLM generates incorrect description of the region regarding its location and action. 
    \item blur only: similar to mask cropped only, using blurred background as the sole visual prompt creates challenges for MLLM in distinguishing boundaries between blurred and clear regions, resulting in inaccurate location identification. Critical action-related details may also be obscured by blurring, leading to incorrect classification of object activities. 
    \item original image with mask-cropped: while adding the original image helps MLLM better understand location and relationships, the lack of explicit region guidance causes MLLM to be distracted by irrelevant regions outside the mask cropped area.
    \item mask cropped with mask contour overlay: adding contour helps MLLM focus on the target region's boundaries, but the choice of overlay color can inadvertently influence MLLM's perception of the region's visual attributes. Attempts to show contours without color overlay (Fig~\ref{fig:overlay-only}) often result in ambiguous or confusing visual prompts, particularly for intricate shapes or overlapping regions if the contour is a non-convex shape.
    \item bounding box with mask contour overlay: while both elements help localize the target region, their overlay colors can affect MLLM's understanding. Even when explicitly prompted to focus on either the bounding box or contour region, both colors influence MLLM's perception of visual attributes, leading to inconsistent descriptions.
    \item bounding box with mask-cropped (\ourmethod): This combination achieves the best balance - the bounding box provides spatial context and relationship guidance, while the mask-cropped image offers detailed visual attributes without color interference. By instructing MLLM to focus on the mask-cropped region while using the bounding box for context, we avoid noise from overlay colors while maintaining accurate spatial understanding.
\end{itemize}

\section{\ourdataset~Data Preparation}
\label{sec:dataset}
\subsection{Image Data}
\begin{figure*}
    \centering
    \includegraphics[width=0.95\linewidth]{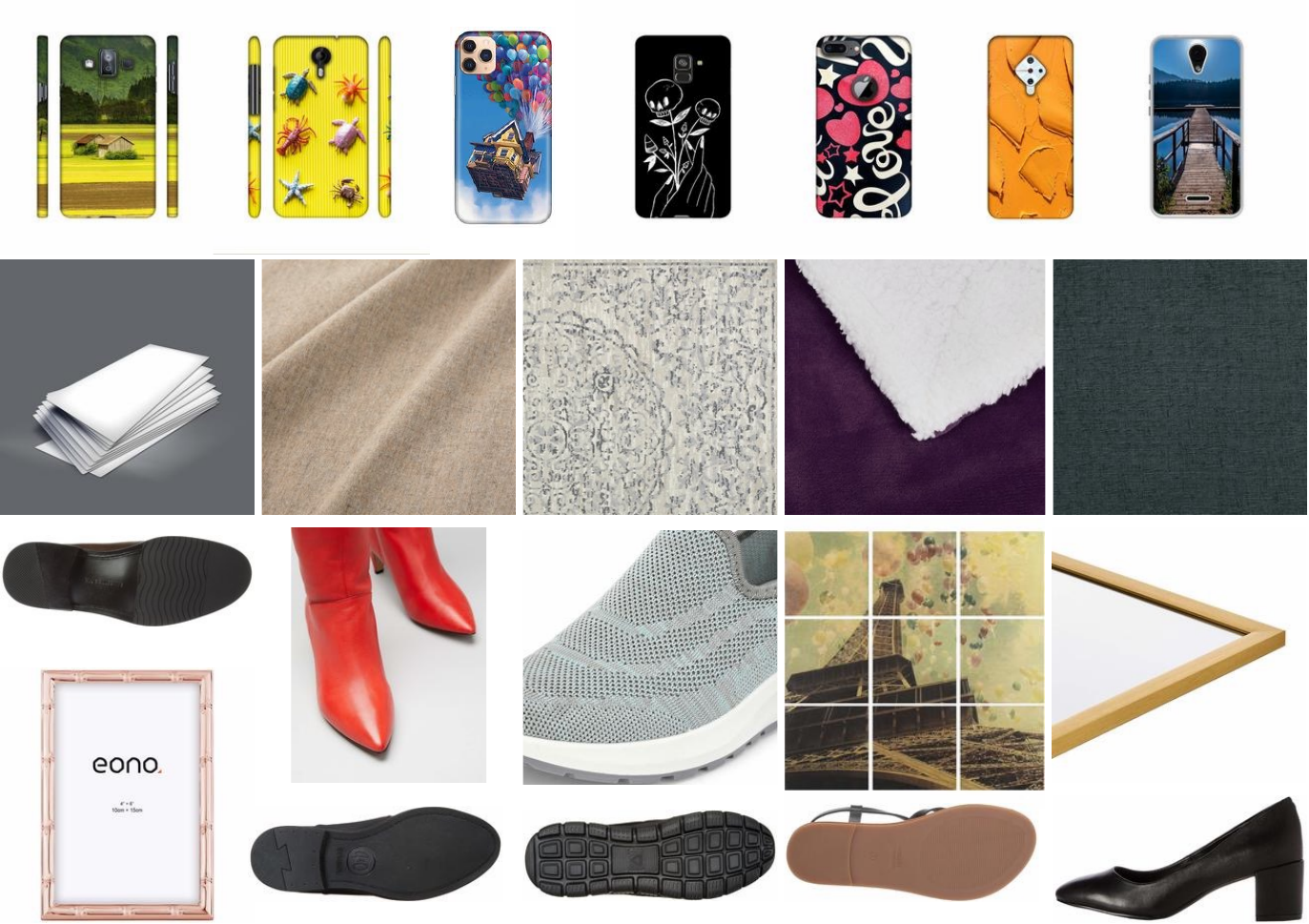}
    \caption{Images excluded from the ABO dataset typically lack meaningful or referrable parts. Row 1 shows phone related items that primarily consist of phone cases displaying only the back view of phones. Row 2 features images solely showing product textures or materials that fill the entire frame. Images from the footwear and picture frames categories in row 3 \& 4 are commonly presented against plain white backgrounds without distinct parts for grounding.}
    \label{fig:abo-remove}
\end{figure*}

Our dataset builds upon image data from ABO~\cite{collins2022abo}, a dataset collected from worldwide Amazon.com product listings, including their metadata, images, and 3D models. ABO encompasses 147,702 product listings across 576 product types from various Amazon-owned stores and websites (e.g., Amazon, PrimeNow, WholeFoods). Each listing is uniquely identified by an item ID and contains structured metadata from its public webpage, including product specifications such as type, material, color, and dimensions, along with associated media. The dataset contains 398,212 high-resolution catalog images in total. However, to better highlight product properties, we excluded \iccv{images from} 11 categories: phone-related items (phone accessories, cellular phone cases, cellular phones, phones, wireless locked phones), footwear (shoes, shoe inserts, technical sport shoes, boots, sandals), and picture frames. \iccv{Most images from these categories have no meaningful or interesting groundable/referrable parts, as shown in Fig ~\ref{fig:abo-remove}.}
We also selected only the main image of each product, as additional images often show material details or close-up views. As results, \ourdataset~contains 2,482 high-resolution catalog images spanning 565 product types.
\subsection{Referring Expression Generation}
The referring expressions in \ourdataset~were derived from product metadata, specifically the bulletpoint descriptions that accompany each product listing in ABO. These bulletpoints typically contain detailed information about product features, materials, and functionalities. We processed these descriptions through MLLM, instructing it to generate 2-3 referring expressions per product. Prompt for instruction is following: \\\\
\texttt{Here is an image of a product. These are the product descriptions for it:  \{bulletpoints\}. Please analyze the descriptions and list 2-3 most important features or functionality. Return key words only without any starting or ending statements. Do not include dimension or assembly information. Each feature should be informative. If you cannot extract any relevant product features from both the image and description, return 'N/A'.} \\\\
To ensure quality and visual grounding, we manually filtered out expressions that is 'N/A' and could not be reliably mapped to specific regions in the product images. \iccv{ We also manually reviewed all generated expressions to ensure the dataset's quality. All manual processing was completed by 4 evaluators. Each evaluator was required to review all the image-expression pairs and judge each expression as either "good" or "bad." To quantify inter-annotator agreement, we employed Fleiss' Kappa~\cite{fleiss1971measuring}, which is suitable for measuring agreement among multiple raters beyond what would be expected by chance. For expressions with low agreement among evaluators (such as 2-2 splits), we either modified the expression manually or removed it from the dataset entirely. The final dataset consists only of expressions that received strong majority approval (3-1 or 4-0 votes) and demonstrated clear visual grounding in the product images. }This rigorous curation process yielded 2,989 referring expressions, each targeting part-level regions and describing specific materials, features, functionalities, or packaging elements. 
\iccv{
\subsection{Mask Annotation}
Our annotation process leverages SAM~\cite{kirillov2023segment} to achieve efficient and accurate region segmentation. The annotation workflow consists of two stages: automatic segmentation and manual refinement. In the first stage, we utilize SAM's automatic mode to generate a comprehensive set of candidate segmentation masks for each image. GT regions that correspond to our referring expressions are then selected from these candidates. For regions that SAM failed to identify automatically, we proceed to the second stage where we manually annotate them using SAM's interactive mode with point supervision. This semi-automated approach significantly streamlines the annotation process while ensuring precise region segmentation for our dataset. }

\iccv{Similar to the evaluation of expressions, we also conducted quality assessment for the segmentation annotations. The same panel of 4 evaluators reviewed each segmented region and classified them as either "good" or "bad" based on their accuracy and alignment with the corresponding expressions. We applied Fleiss' Kappa~\cite{fleiss1971measuring} to measure inter-annotator agreement for these segmentation evaluations as well. Regions with low agreement scores were flagged for re-annotation using more precise point supervision in SAM's interactive mode. Only segmentations that received strong majority approval (3-1 or 4-0 votes) were retained in the final dataset, ensuring that our ground truth regions accurately represent the visual elements referenced in the expressions. }
\section{Quantitative Results}
\label{sec:quan}
\iccv{To ensure statistical robustness and account for potential variability in \ourmethod's performance, especially for the components involving LLM generation (reference text, candidate text, and similarity analysis), we conducted experiments with our approach 8 separate times and report the averaged results in both the main paper and supplementary materials. }
\subsection{CLIP as RNN}
We present quantitative results of CLIP as RNN, the current SOTA zero-shot method, on both ReasonSeg and \ourdataset~in Table~\ref{tab:quan-car}. 
\begin{table}[h]
\caption{Quantitative results of CLIP as RNN~\cite{sun2024clip}, with \ourmethod's results shown in parentheses for comparison.}
\vspace{-0.5em}
\label{tab:quan-car}
\centering
\footnotesize
\begin{tabular}{llcc}
\toprule
\multirow{2}{*}{Dataset} & & \multicolumn{2}{c}{test} \\\cmidrule{3-4}
&  & \multicolumn{1}{c}{gIoU (\textbf{ours})}  & \multicolumn{1}{c}{cIoU (\textbf{ours})}\\
\midrule
ReasonSeg\cite{lai2024lisa} & & 35.2 \textbf{(74.6)} & 26.4 \textbf{(72.5)} \\
\ourdataset &  & 24.4 \textbf{(78.2)} & 15.7 \textbf{(72.4)}\\
\bottomrule
\end{tabular}
\end{table}

\subsection{Part-only RES benchmark}
We further evaluate the performance of \ourmethod{} and competing methods on UniRES~\cite{wang2024unveiling}, which contains a subset RefCOCOm for part-level RES. Table~\ref{tab:refcocom} shows the quantitative results on {\em part-only\/} RefCOCOm. Since the code for UniRES~\cite{wang2024unveiling} is not publicly available, we directly compare performances using the mIoUs reported in their paper. Although UniRES is claimed to be a zero-shot method, it is pre-trained on their proposed MRES-32M dataset, which is closed source. Our method significantly outperforms the training-free zero-shot CaR, and generally outforms the supervised UniRES and LISA, {\em even though\/} they were both pre-trained on related tasks. GLaMM is the same and is slightly ahead of ours, but this is attributable to its additional fine-tuning on their proposed GranD dataset.
\begin{table}[h]
\caption{Quantitative results on RefCOCOm \textbf{Part-only} set.}
\vspace{-0.5em}
\label{tab:refcocom}
\centering
\footnotesize
\begin{tabular}{lccc}
\toprule
Method & val & testA & testB \\
\midrule
\emph{supervised / pre-trained}\\
UniRES~\cite{wang2024unveiling} & 19.6 & 16.4 & 25.2 \\
LISA~\cite{lai2024lisa} & 21.2 & 19.1 & 27.4\\
\rowcolor{red!10!white}GLaMM~\cite{rasheed2024glamm} & 30.0 & 27.2 & 31.8\\
\midrule
\emph{training-free zero-shot}\\
CaR~\cite{sun2024clip} & 10.9 & 10.6 & 10.9\\
\rowcolor{blue!10!white}RESAnything & 27.6 & 26.5 & 25.8\\
\bottomrule
\end{tabular}
\end{table}
\subsection{Multi-object GRES benchmarks}
Although \ourmethod{} is not specifically designed for multi-object RES task, it still effectively handles these cases through the grouping and selection algorithm, demonstrating the generalization on these tasks. Table~\ref{tab:rebuttal-grefcoco} reports qualitative comparison on a GRES benchmark, g-RefCOCO~\cite{liu2023gres}. Among the methods, only GRES is trained on g-RefCOCO. RESAnything achieves comparable results as LISA and GLaMM, while significantly outperforming the training-free zero-shot method CaR. 
\begin{table}[h]
    \centering
    \caption{Results on gRefCOCO (cIoU).}
    \vspace{-0.5em}
    \label{tab:rebuttal-grefcoco}
    \footnotesize
    \setlength{\tabcolsep}{3pt}
    \begin{tabular}{lccc}
    \toprule
    Method & val & testA & testB \\
    \midrule
    \emph{pre-trained on vanilla RES tasks} \\
    LISA~\cite{lai2024lisa} & 48.4 & 45.1 & 46.3\\
    GLaMM~\cite{rasheed2024glamm} & 46.2 & 46.7 & 47.2\\
    \midrule
    \emph{supervised (trained on gRefCOCO)} \\
    \rowcolor{red!10!white}GRES~\cite{liu2023gres} & 62.4 & 69.3 & 59.9 \\
    \midrule
    \emph{training-free zero-shot} \\
    CaR~\cite{sun2024clip} & 25.6 & 22.0 & 21.5\\
    \rowcolor{blue!10!white}RESAnything & 52.7 & 46.2 & 46.3 \\
    \bottomrule
    \end{tabular}
\end{table}

We conducted additional evaluations of our method against competing methods on R-RefCOCO~\cite{wu2024towards} and RefZOM~\cite{hu2023beyond}. Images in both datasets are extracted from the RefCOCO, with additional multi-object referring expressions. Table~\ref{tab:rebuttal-rrefcoco} shows the quantitative results on both benchmarks. Both RefSegformer~\cite{wu2024towards} and DMMI~\cite{hu2023beyond} are fully supervised method trained on the training set of R-RefCOCO and RefZOM separately. LISA and GLaMM also pre-trained on image data from COCO, which serves as the based of both benchmarks. Our method reasonaly underperformed against supervised methods that were explicitly exposed to the training set, but still outperforms the SOTA training-free zero-shot baseline. 

\begin{table}[h]
\centering
\caption{Results on R-RefCOCO and RefZOM(mIoU)}
\vspace{-0.5em}
\label{tab:rebuttal-rrefcoco}
\footnotesize
\setlength{\tabcolsep}{0.5pt}
\begin{tabular}{lccc}
\toprule
Method & R-RefCOCO & &  RefZOM  \\
\midrule
\emph{supervised (trained on training set)} \\
RefSegformer~\cite{wu2024towards} & 68.8 & & -\\
DMMI~\cite{hu2023beyond} & - & & \cellcolor{red!10!white}68.2 \\
\midrule
\emph{pre-trained} \\
LISA~\cite{lai2024lisa} & 71.1 & & 45.0\\
GLaMM~\cite{rasheed2024glamm} & \cellcolor{red!10!white}72.1 & & 47.4\\
\midrule
\emph{training-free zero-shot} \\
CaR~\cite{sun2024clip} & 30.2 & & 25.7\\
\rowcolor{blue!10!white}RESAnything & 61.2 & & 40.3\\
\bottomrule
\end{tabular}
\end{table}

\subsection{Runtime Comparison}
\iccv{As stated in the main paper, our method's entire inference process can run efficiently on a single NVIDIA 24GB 4090 GPU. For a fair comparison, we measured the execution times of all competing methods on the same hardware. The average per-image processing time was evaluated on the ReasonSeg test set, with detailed results provided in Table~\ref{tab:runtime}. While our main results in the main paper were conducted using 8 V100 GPUs for running multiple experiments in parallel during development, we optimized our method's runtime for comparison experiments. These optimizations include: 1) utilizing the bfloat16 data format for the LLM, which is not supported on V100; 2) enabling flash attention for more efficient transformer operations; 3) implementing batch generation for LLM outputs rather than sequential processing of each reference and candidate text; and 4) employing batch computation for CLIP similarity scores.}


\begin{table}[h]
\caption{Runtime comparison.}
\vspace{-0.5em}
\label{tab:runtime}
\centering
\footnotesize
\begin{tabular}{llc}
\toprule
Method & & Time/image (s)\\
\midrule
CaR & & 5.3\\
LISA & & 7.0\\
GLaMM & & 8.6\\
\ourmethod-Qwen 2-VL & & 12.1 \\
\bottomrule
\end{tabular}
\end{table}

\section{Qualitative Results}
\label{sec:qual}
\begin{figure*}
    \centering
    \includegraphics[width=0.95\linewidth]{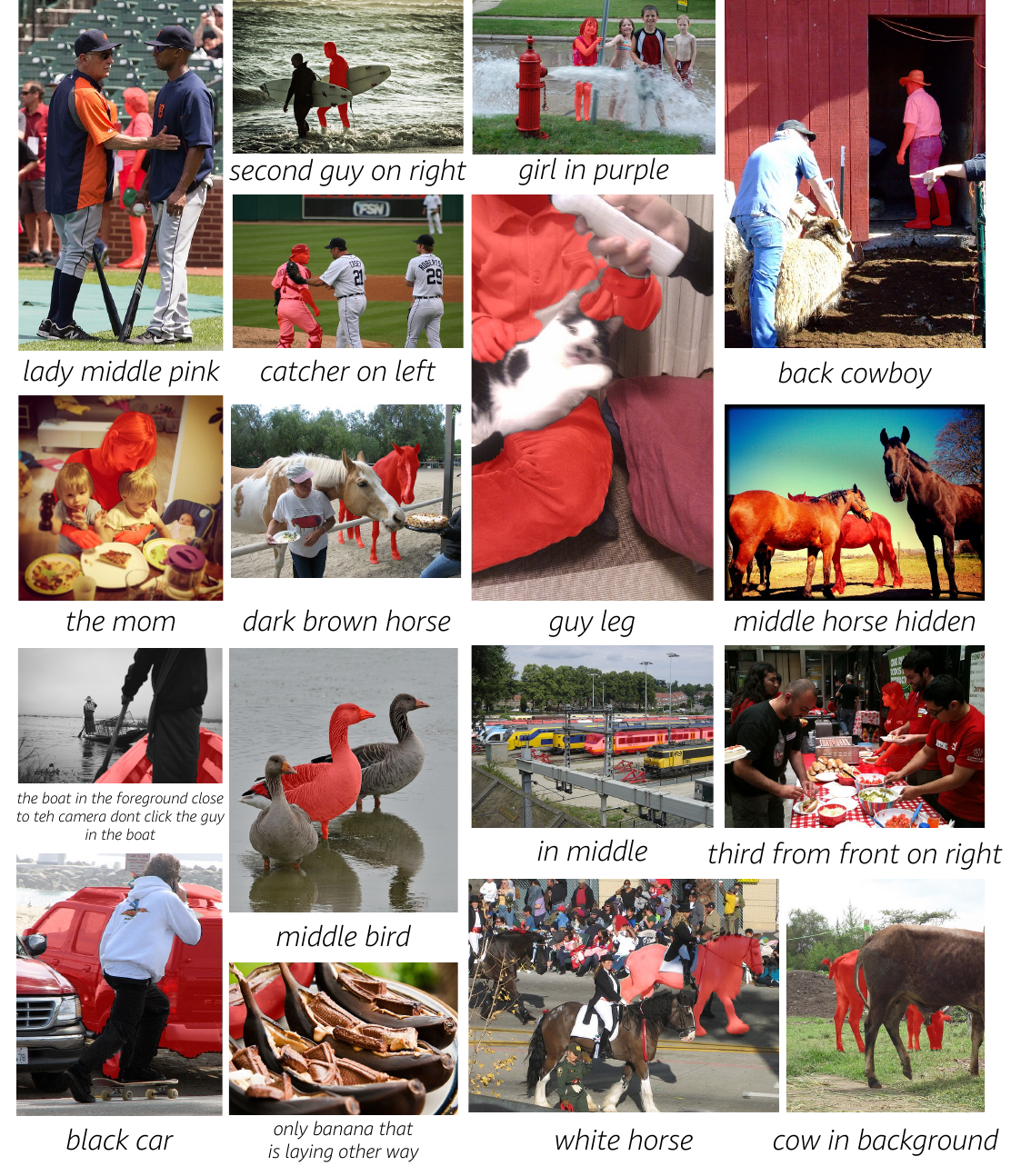}
    \caption{\ourmethod{} can handle occlusion cases by grouping and selection cases. Results from RefCOCO. }
    \label{fig:occlusion}
\end{figure*}

\begin{figure*}
    \centering
    \includegraphics[width=\linewidth]{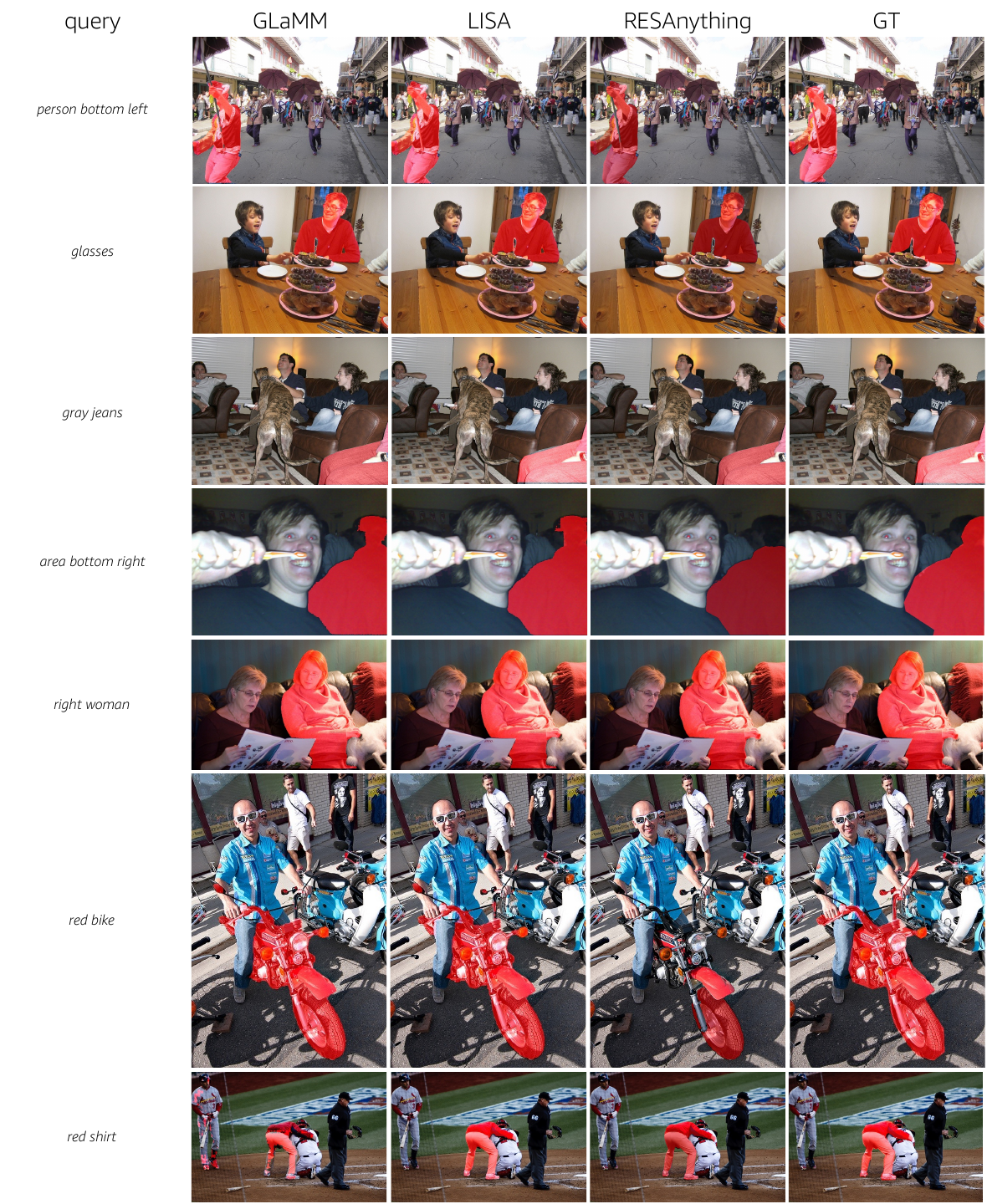}
    \caption{Qualitative results on RefCOCO test A (randomly selected). The ground truth annotations can be problematic - some queries refer only to an object/region while the GT marks an entire person (row 2: ``glasses"; row 4: ``area bottom right", row 7: ``red shirt"). Row 6 shows a failure case of \ourmethod.}
    \label{fig:qual-refcoco-test-a}
\end{figure*}

\begin{figure*}
    \centering
    \includegraphics[width=\linewidth]{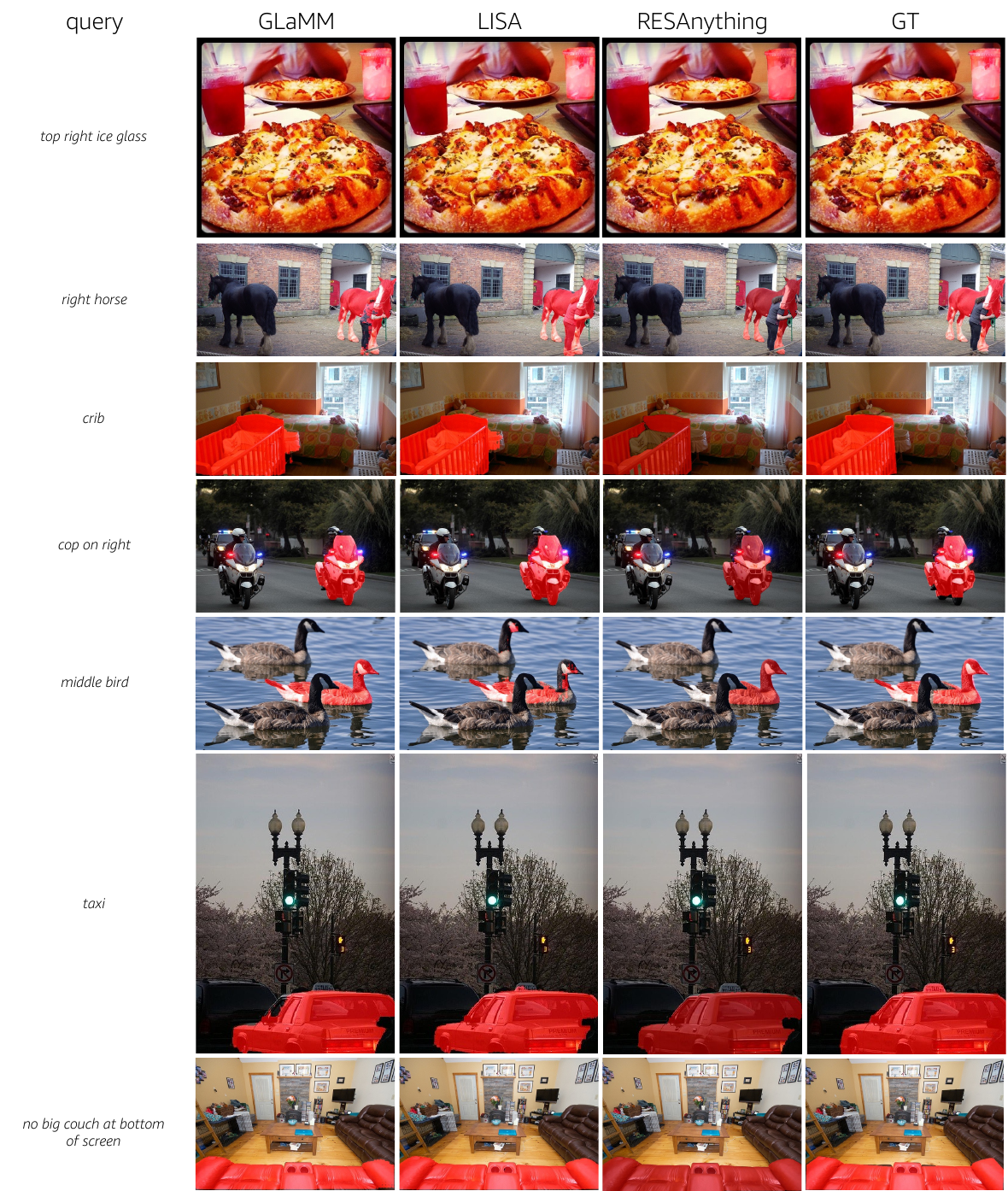}
    \caption{Qualitative results on RefCOCO test B (randomly selected). Despite being unsupervised, \ourmethod~achieves comparable results to supervised methods, particularly excelling at crowded regions (row 2). However, it occasionally misses parts when needing to combine multiple masks (row 3, 5).}
    \label{fig:qual-refcoco-test-b}
\end{figure*}

\begin{figure*}
    \centering
    \includegraphics[width=\linewidth]{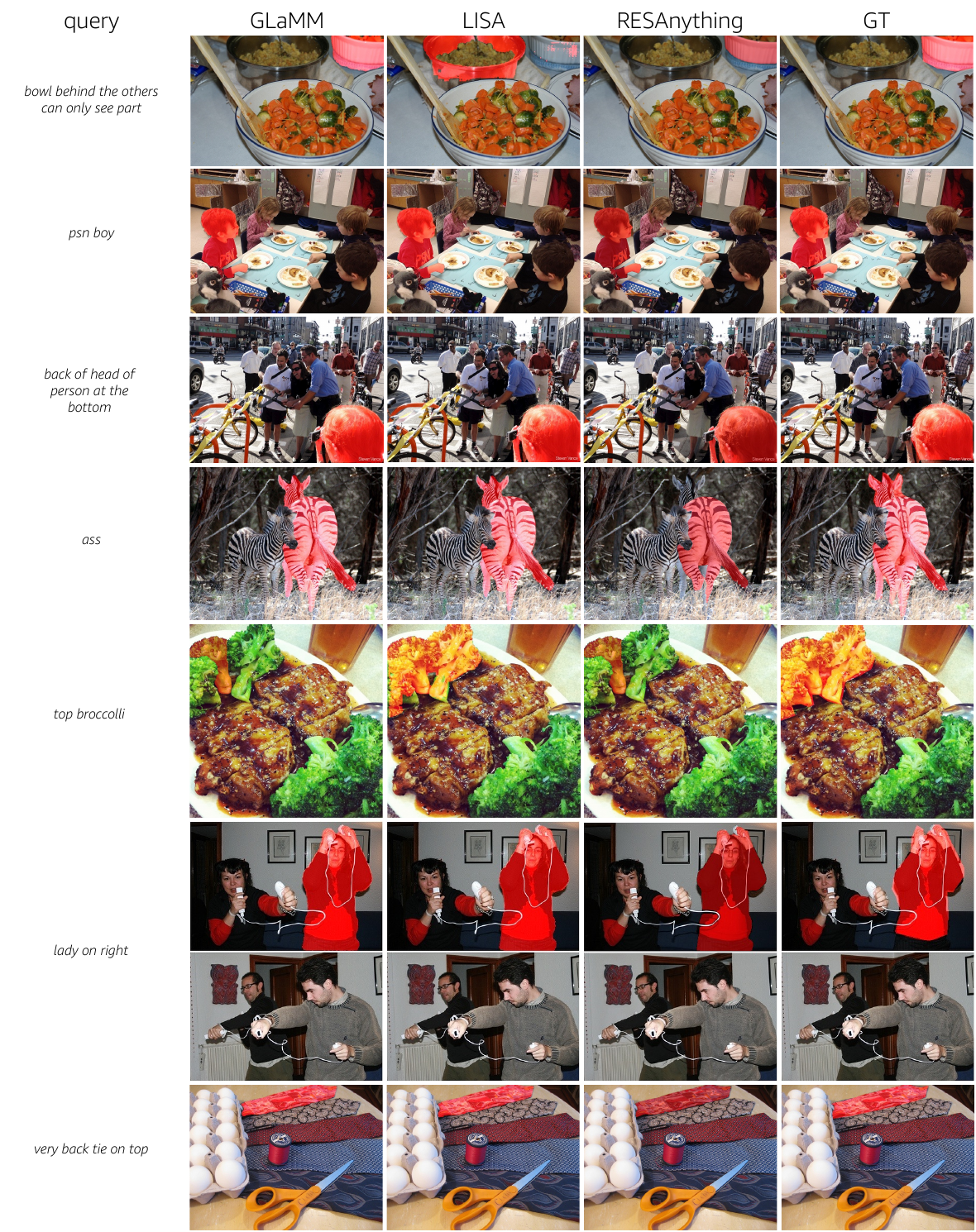}
    \caption{Qualitative results on RefCOCO val (randomly selected). \ourmethod~generates fine-grained segmentation of the input expression (row 1, 4). As mentioned in Fig~\ref{fig:qual-refcoco-test-b}, it may miss parts when combining multiple masks (row 5). }
    \label{fig:qual-refcoco-val}
\end{figure*}

\begin{figure*}
    \centering
    \includegraphics[width=\linewidth]{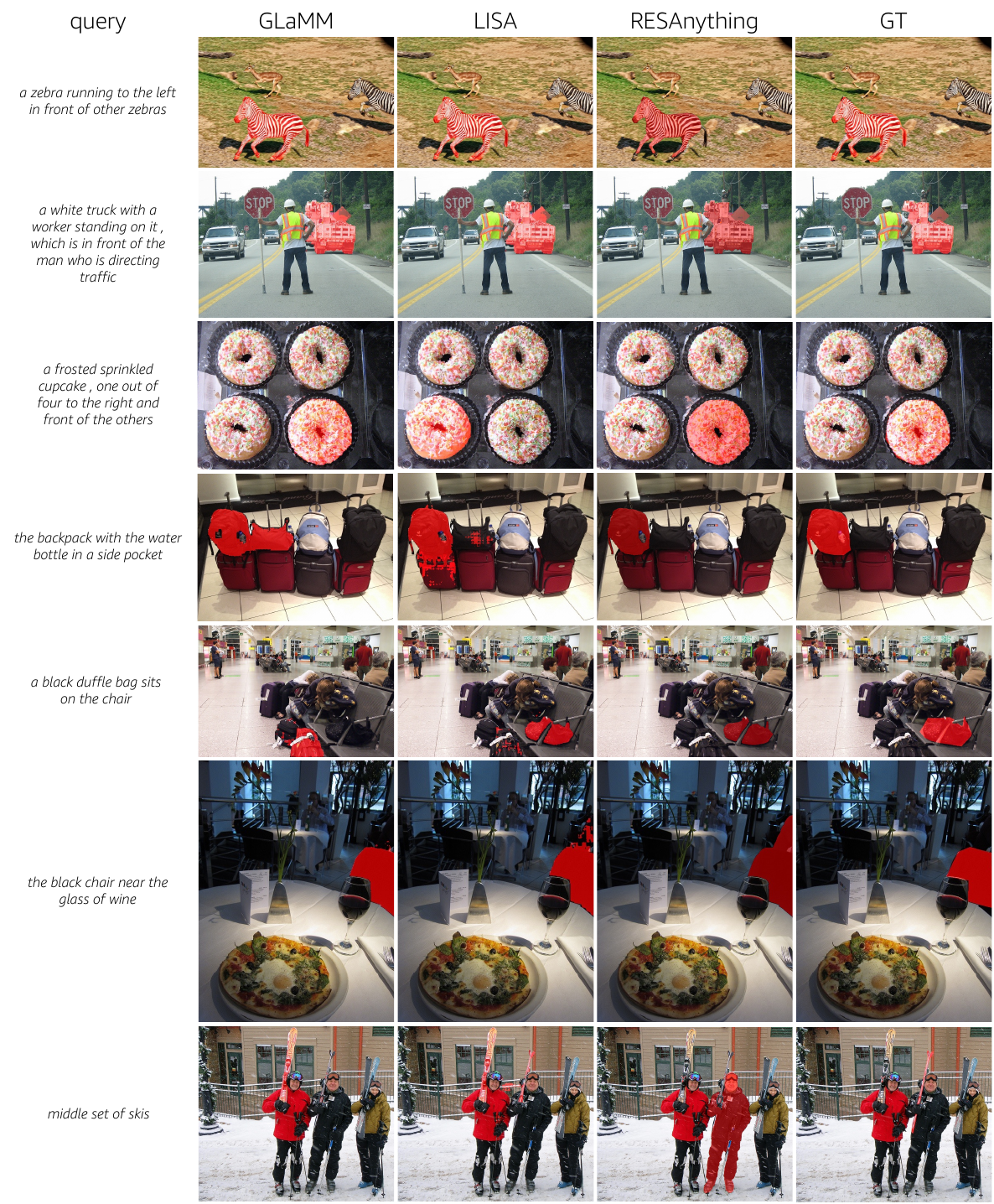}
    \caption{Qualitative results on RefCOCOg val (G) (randomly selected). \ourmethod~generalizes well on mask with hole (row 3, 5), but may suffering from over-segmentation (row 1, 4) or no good candidate found (row 7).}
    \label{fig:qual-refcocogg-val}
\end{figure*}

\begin{figure*}
    \centering
    \includegraphics[width=\linewidth]{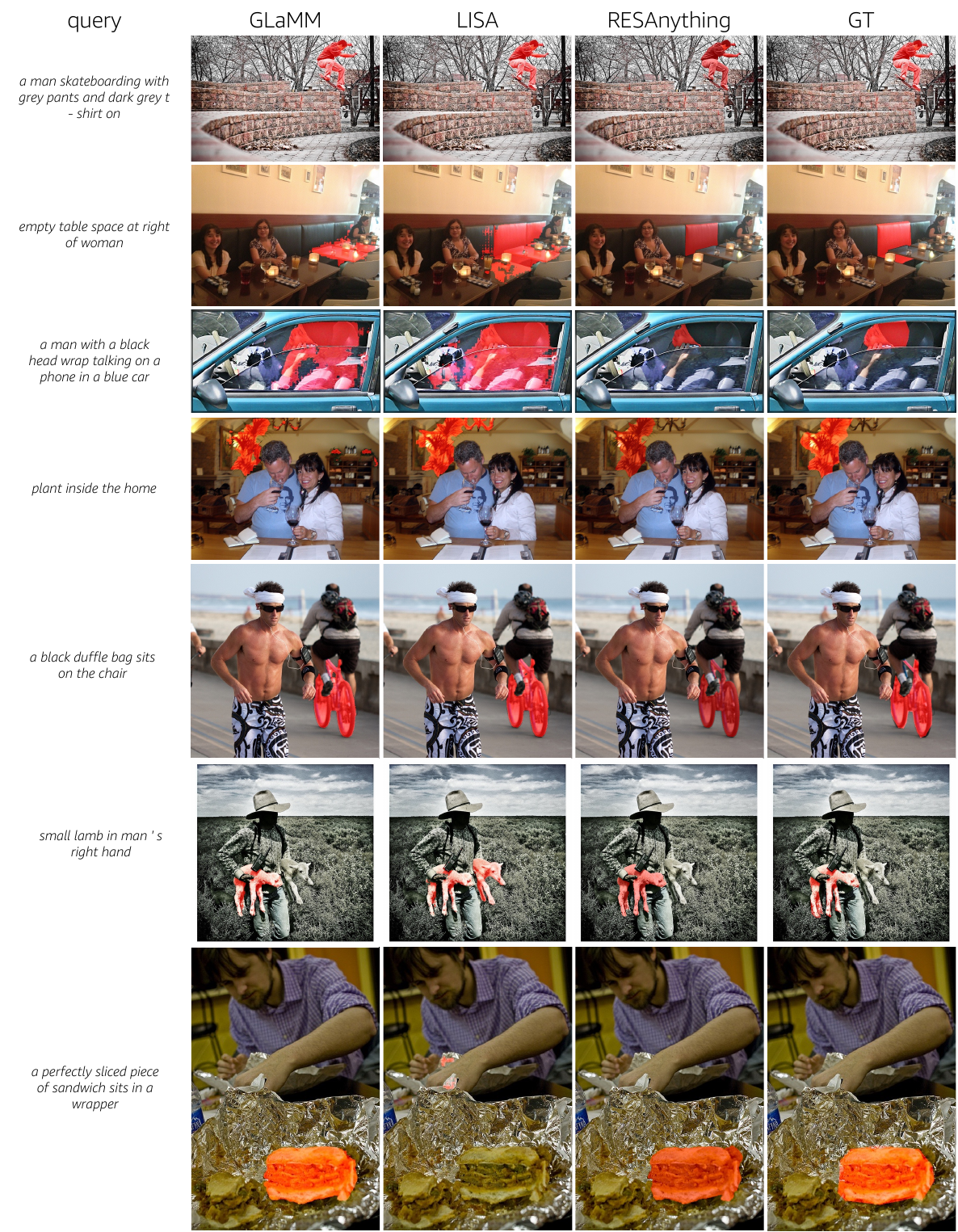}
    \caption{Qualitative results on RefCOCOg val (U) (randomly selected). }
    \label{fig:qual-refcocogu-val}
\end{figure*}

\begin{figure*}
    \centering
    \includegraphics[width=\linewidth]{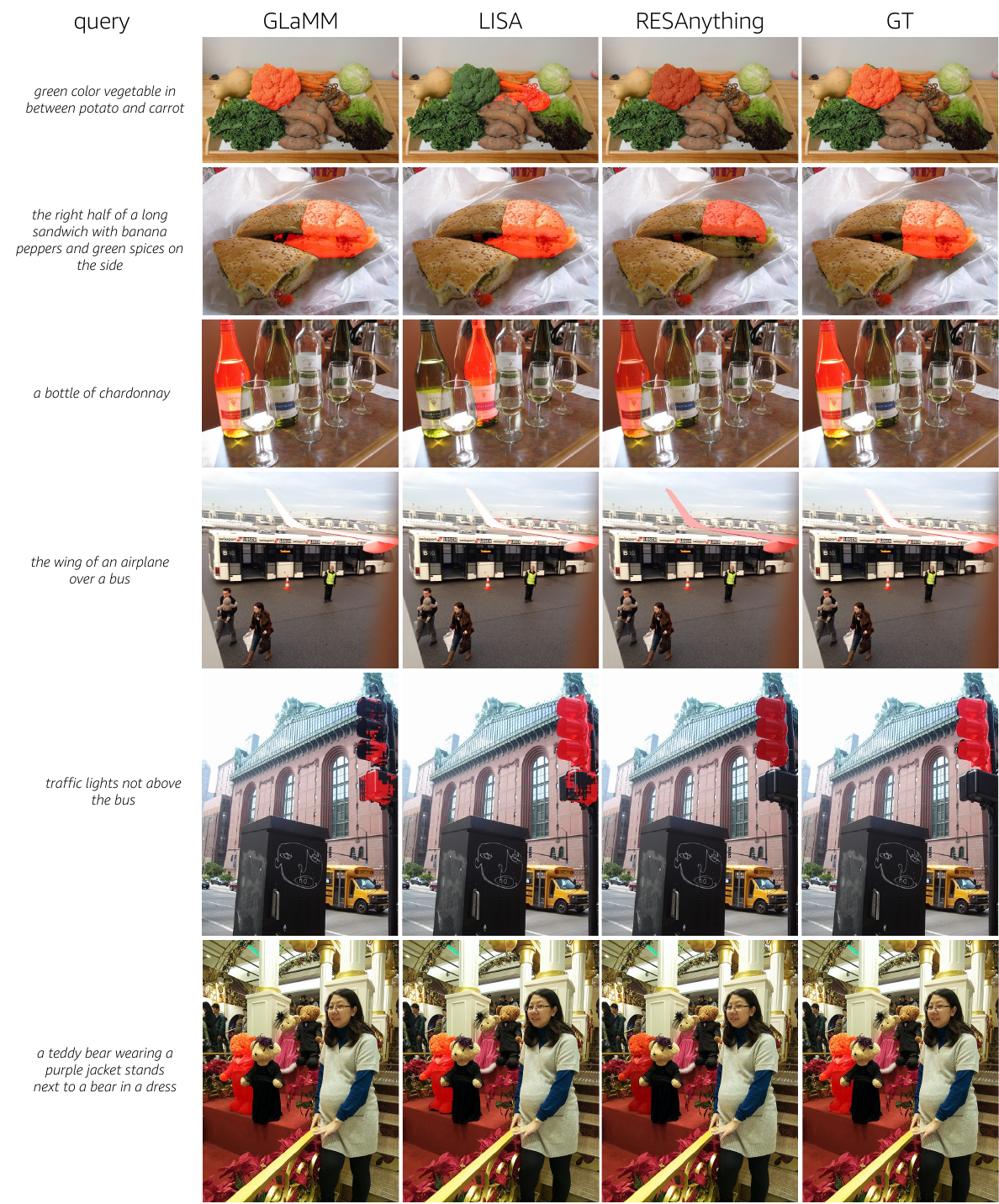}
    \caption{Qualitative results on RefCOCOg test (U) (randomly selected). }
    \label{fig:qual-refcocogu-test}
\end{figure*}

\begin{figure*}
    \centering
    \includegraphics[width=\linewidth]{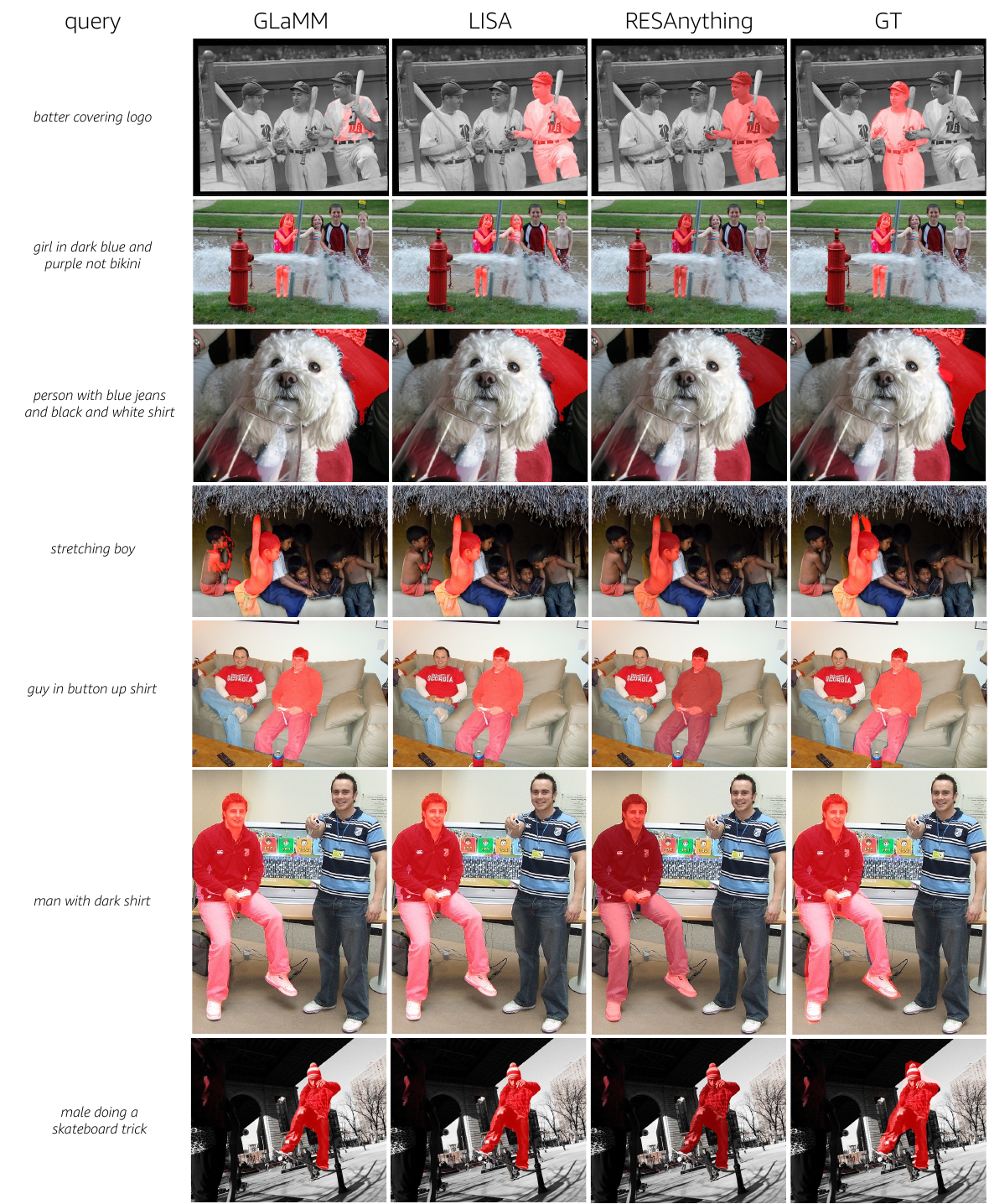}
    \caption{Qualitative results on RefCOCO+ test A (randomly selected). }
    \label{fig:qual-refcocoplus-testa}
\end{figure*}

\begin{figure*}
    \centering
    \includegraphics[width=\linewidth]{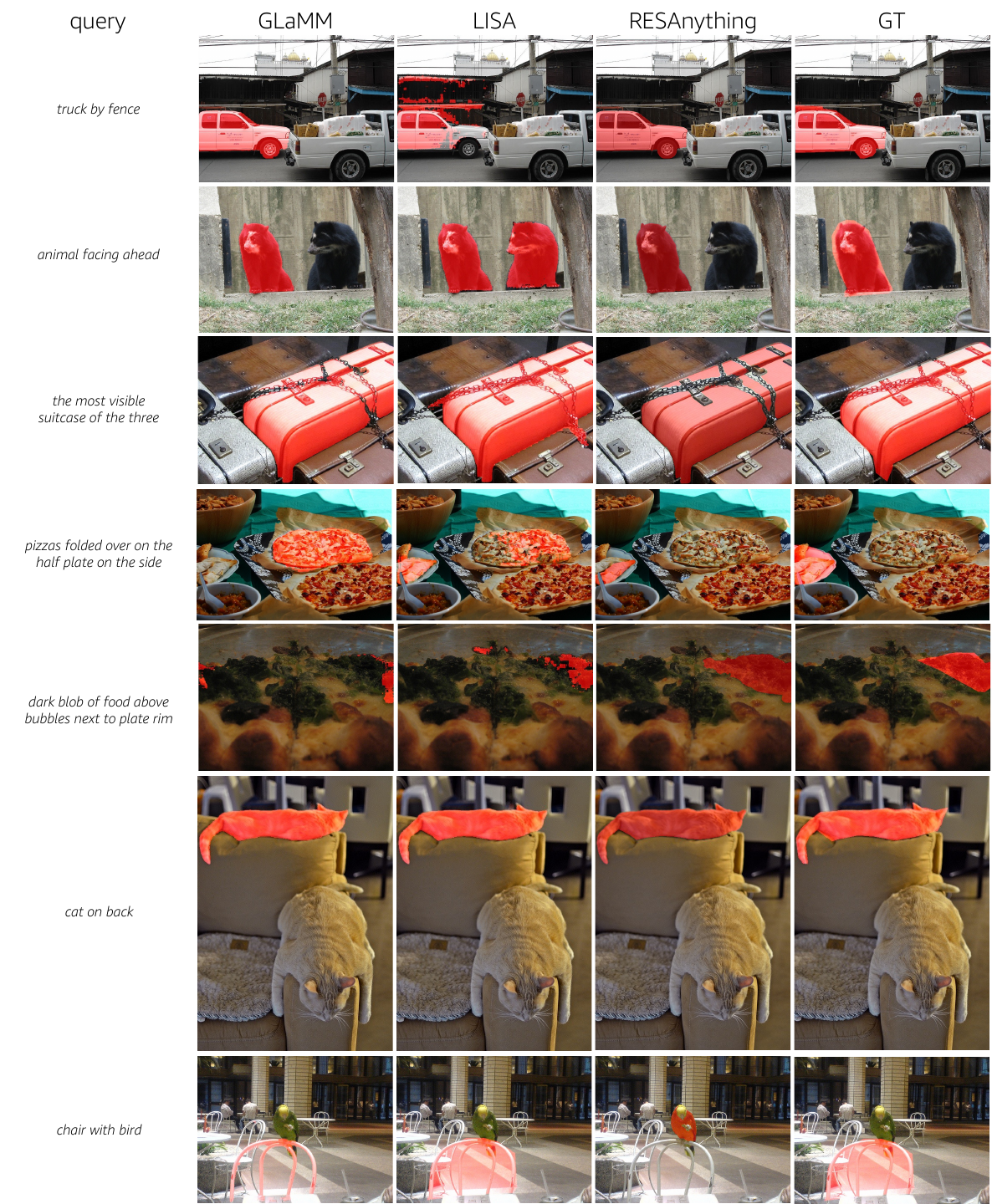}
    \caption{Qualitative results on RefCOCO+ test B (randomly selected). }
    \label{fig:qual-refcocoplus-testb}
\end{figure*}

\begin{figure*}
    \centering
    \includegraphics[width=\linewidth]{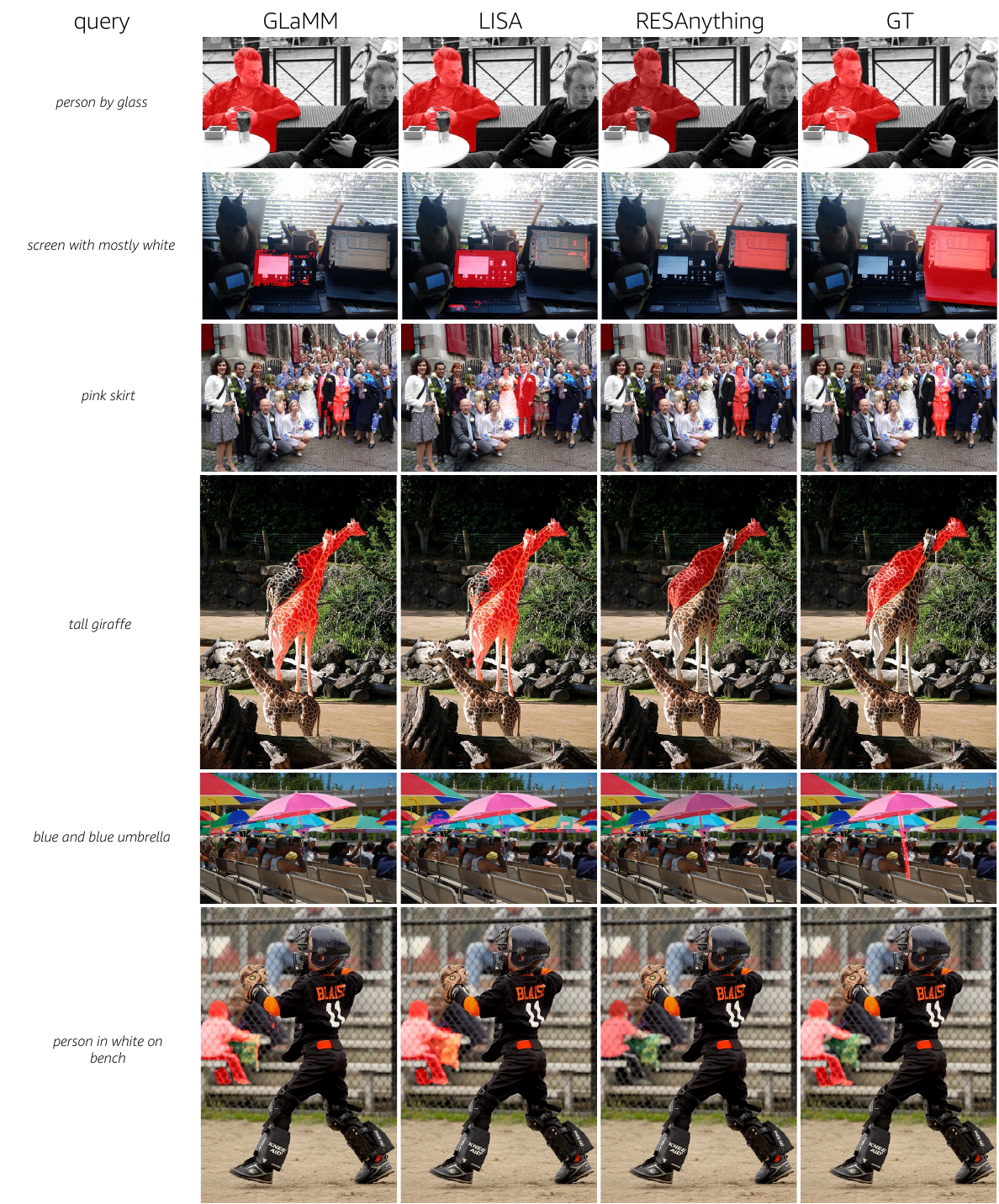}
    \caption{Qualitative results on RefCOCO+ val (randomly selected). }
    \label{fig:qual-refcocoplus-val}
\end{figure*}

\iccv{Firstly, Fig~\ref{fig:qual-refcoco-test-a} --~\ref{fig:qual-refcocoplus-val} show qualitative results on RefCOCO test A, test B, RefCOCOg val (G), val (U), test (U), RefCOCO+ test A, test B, val set separately. \textbf{These examples are randomly selected to provide an unbiased assessment.} \ourmethod~achieves comparable results to supervised methods on vanilla referring segmentation tasks. While our approach effectively handles many occlusion cases, as shown in Fig~\ref{fig:occlusion}, neither our method nor current SOTA approaches can guarantee perfect part detection in every inference. Some failure cases in these results demonstrate challenges in combining parts or handling occlusions. Detailed descriptions and analysis of these failure cases are provided in the individual figure captions. }

\begin{figure*}
    \centering
    \includegraphics[width=\linewidth]{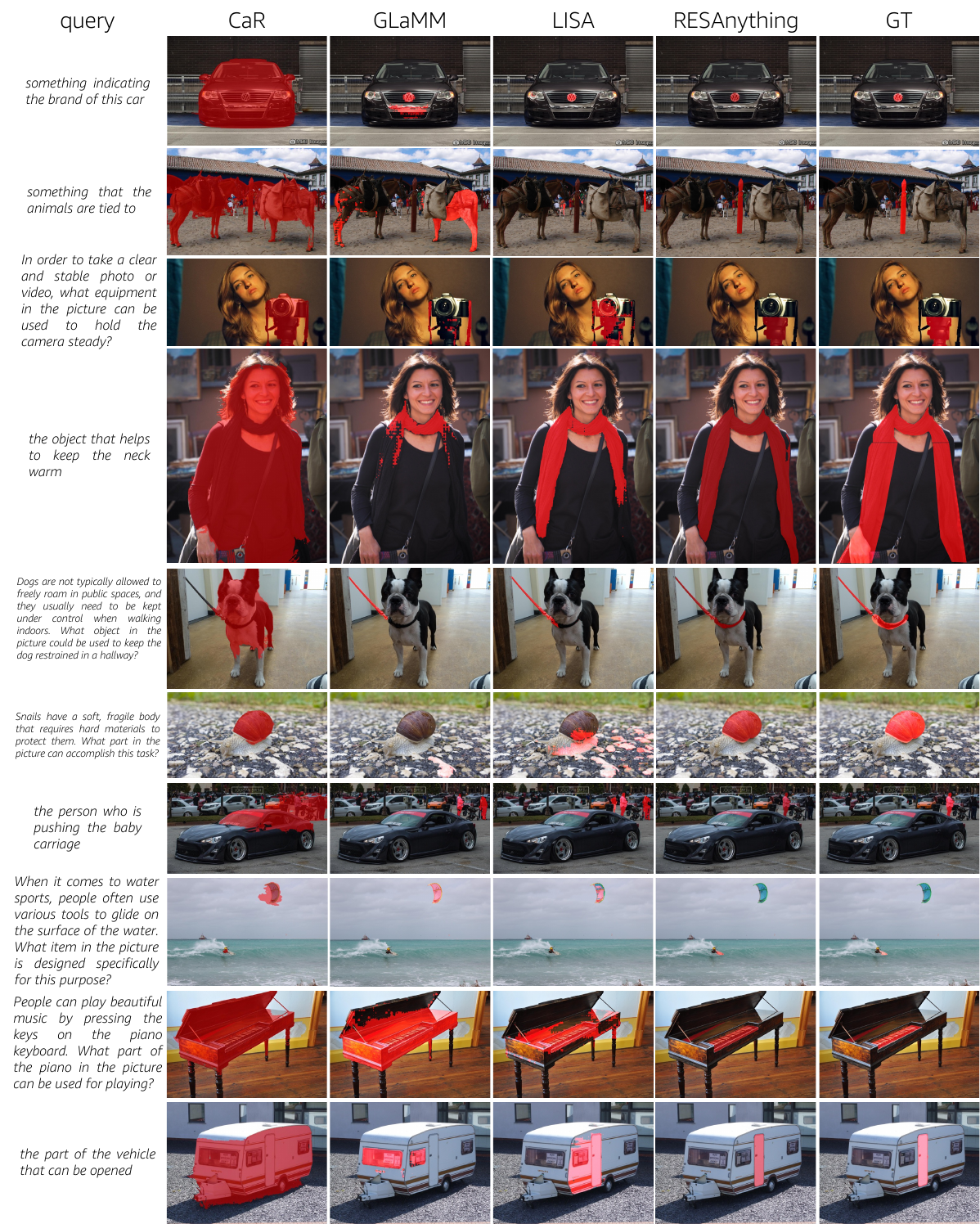}
    \caption{Qualitative results on ReasonSeg (Part 1). \ourmethod~outperforms others in correct localization (row 2, 3, 8, 9), refined segmentation (row 1, 4, 5, 6, 7) and part-level understanding (row 9, 10).}
    \label{fig:qual-reasonseg-1}
\end{figure*}

\begin{figure*}
    \centering
    \includegraphics[width=\linewidth]{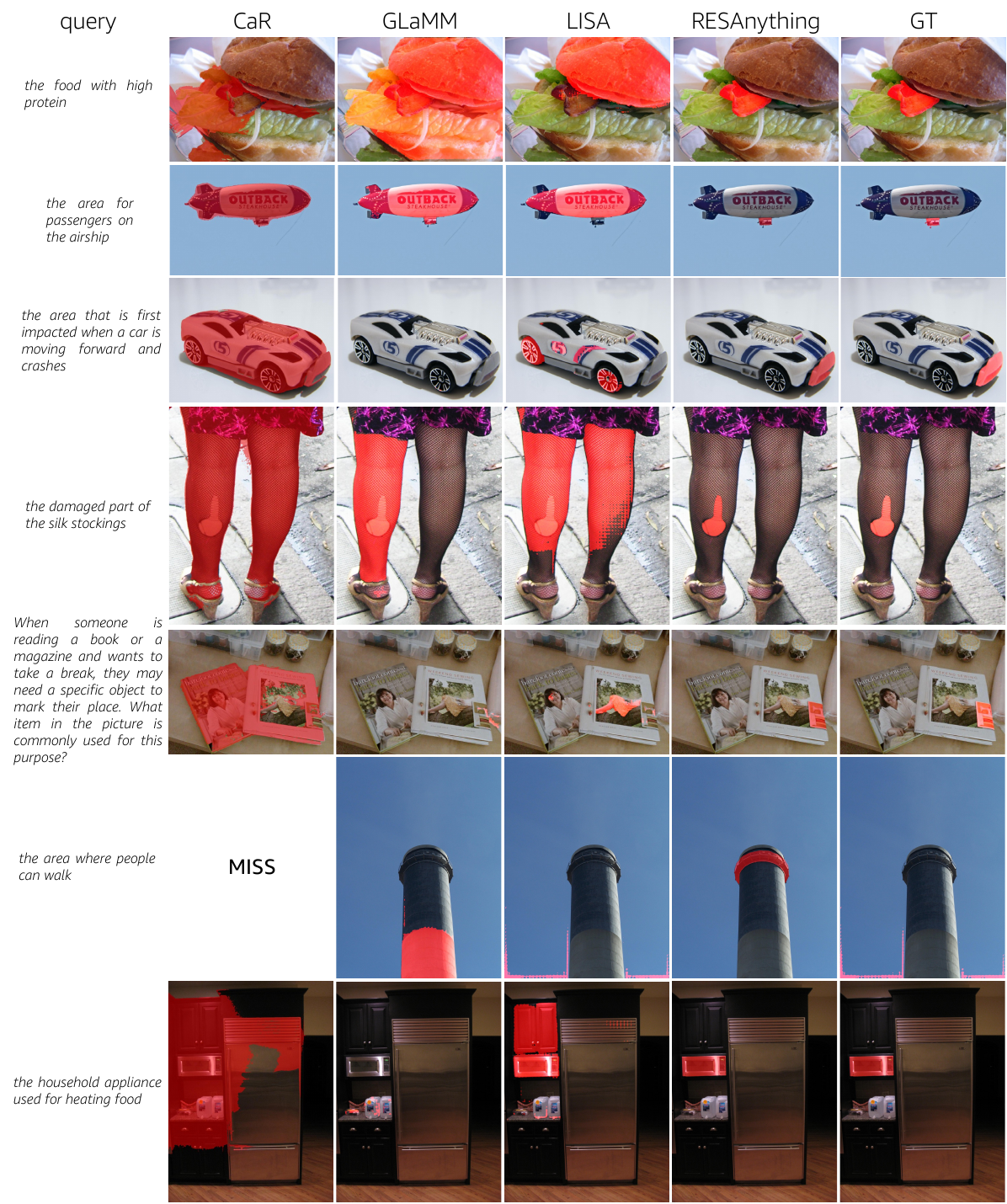}
    \caption{Qualitative results on ReasonSeg (Part 2). ``MISS" indicates that the method is failed to output a segmentation. }
    \label{fig:qual-reasonseg-2}
\end{figure*}

\begin{figure*}
    \centering
    \includegraphics[width=\linewidth]{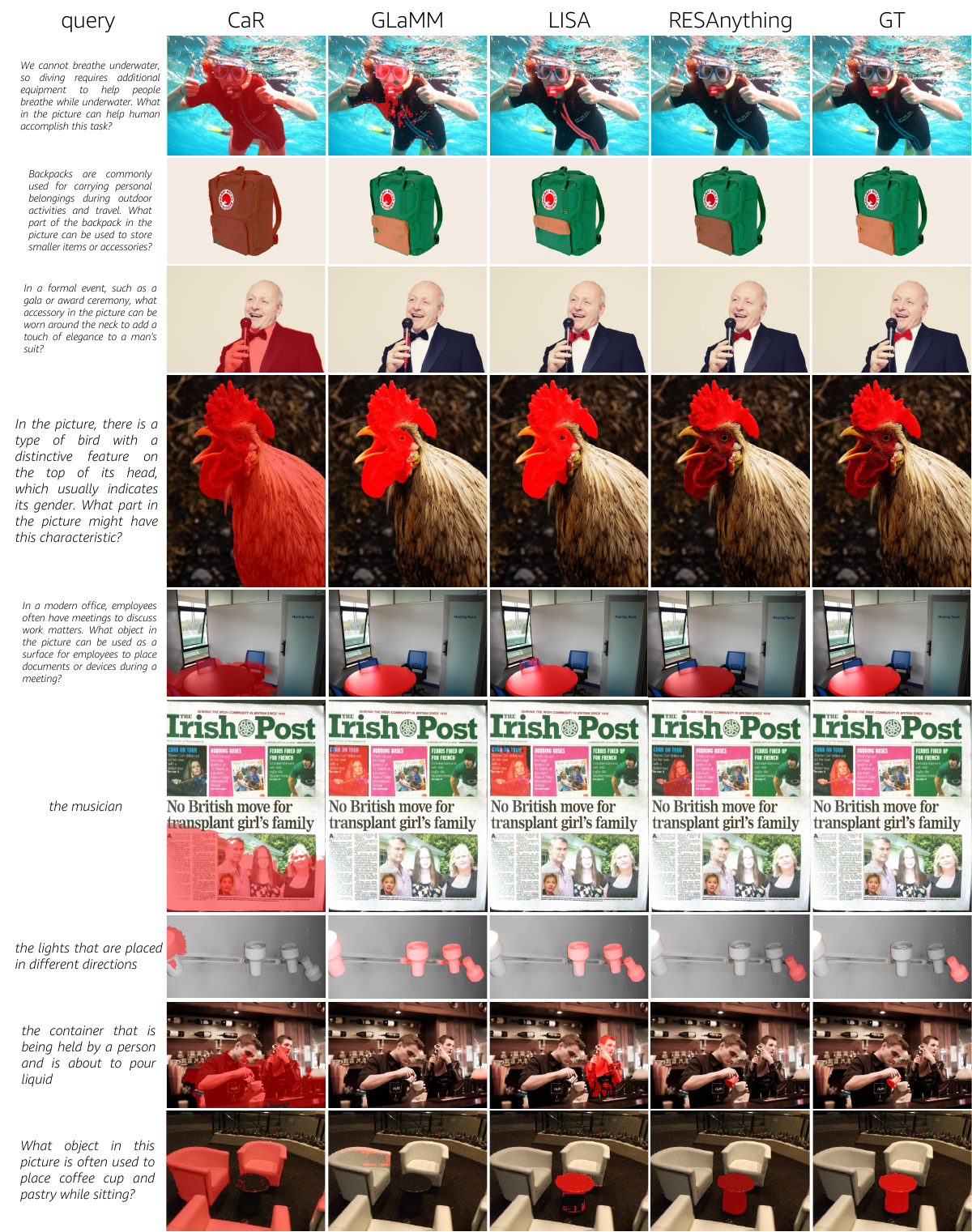}
    \caption{Qualitative results on ReasonSeg (Part 3).}
    \label{fig:qual-reasonseg-3}
\end{figure*}

Secondly, additional qualitative results on ReasonSeg~\cite{lai2024lisa} are shown in Figs.~\ref{fig:qual-reasonseg-1}--\ref{fig:qual-reasonseg-3}, including comparisons with CLIP as RNN (CaR)~\cite{sun2024clip}, the current SOTA zero-shot method. Compared to supervised methods, our Chain-of-Thoughts attribute prompting enables clearer input expression identification and more precise segmentation results. Detailed descriptions and failure cases are provided in the captions.

\begin{figure*}
    \centering
    \includegraphics[width=\linewidth]{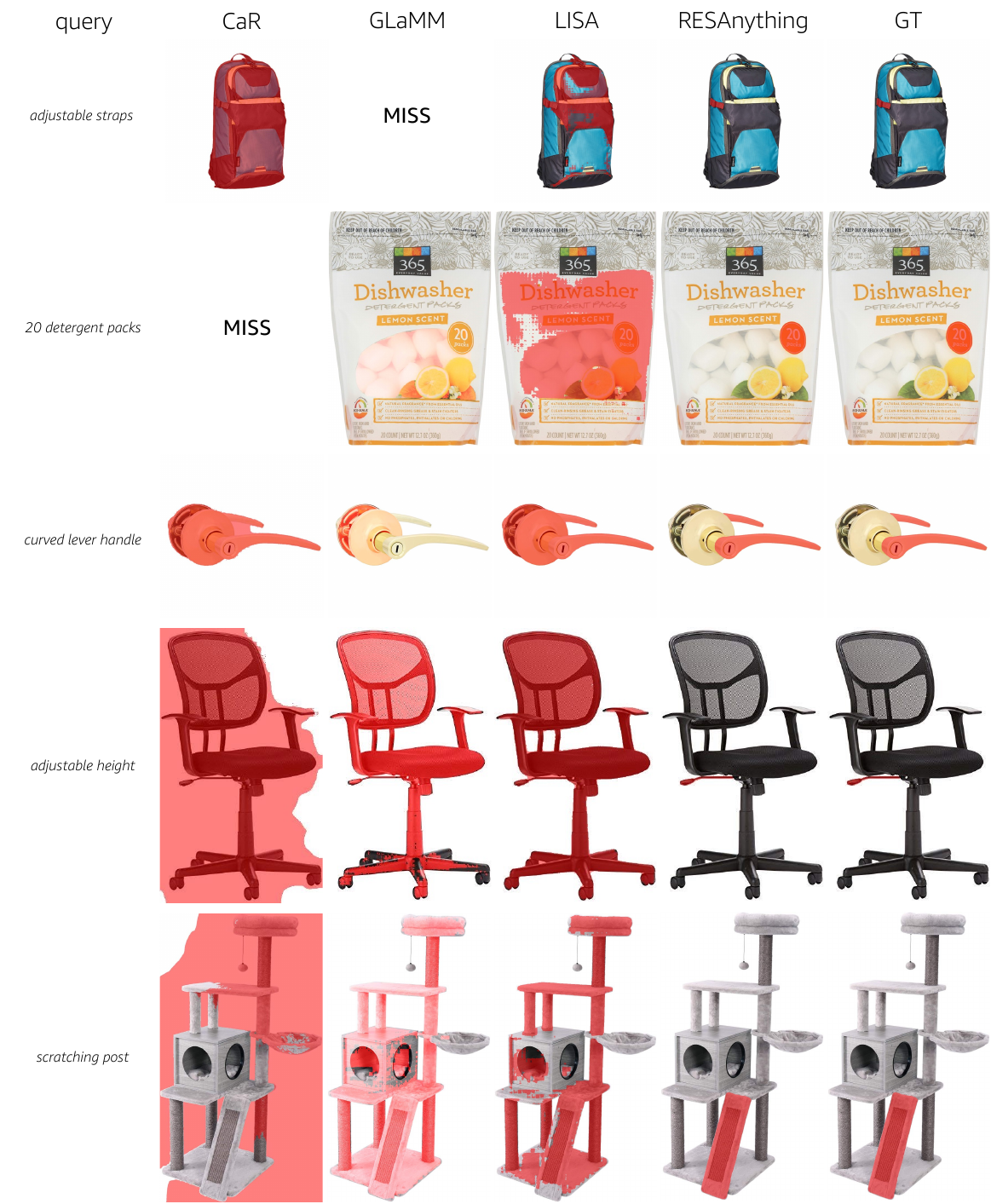}
    \caption{Qualitative results on \ourdataset~(Part 1). }
    \label{fig:qual-abo-1}
\end{figure*}

\begin{figure*}
    \centering
    \includegraphics[width=\linewidth]{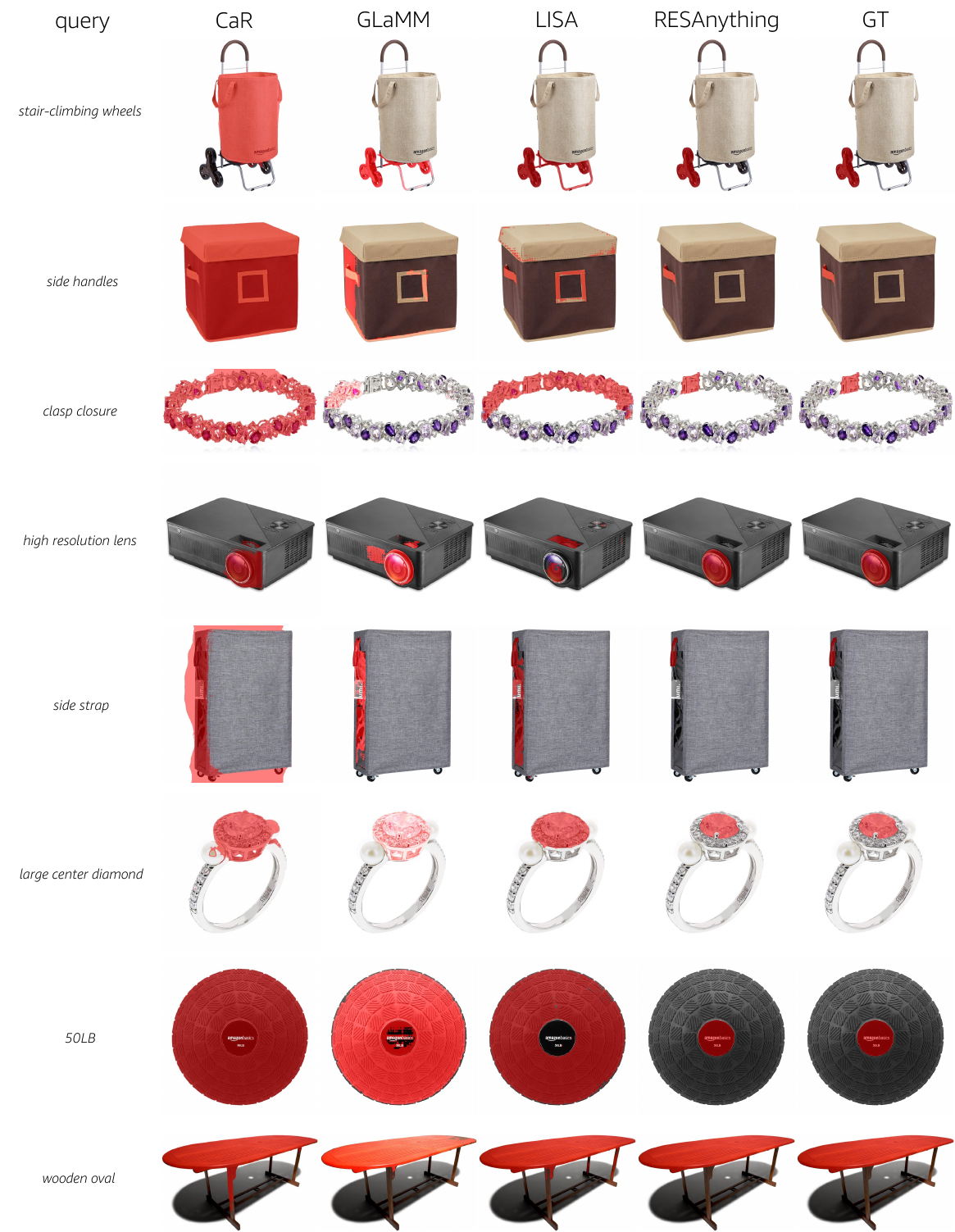}
    \caption{Qualitative results on \ourdataset~(Part 2). }
    \label{fig:qual-abo-2}
\end{figure*}

\begin{figure*}
    \centering
    \includegraphics[width=\linewidth]{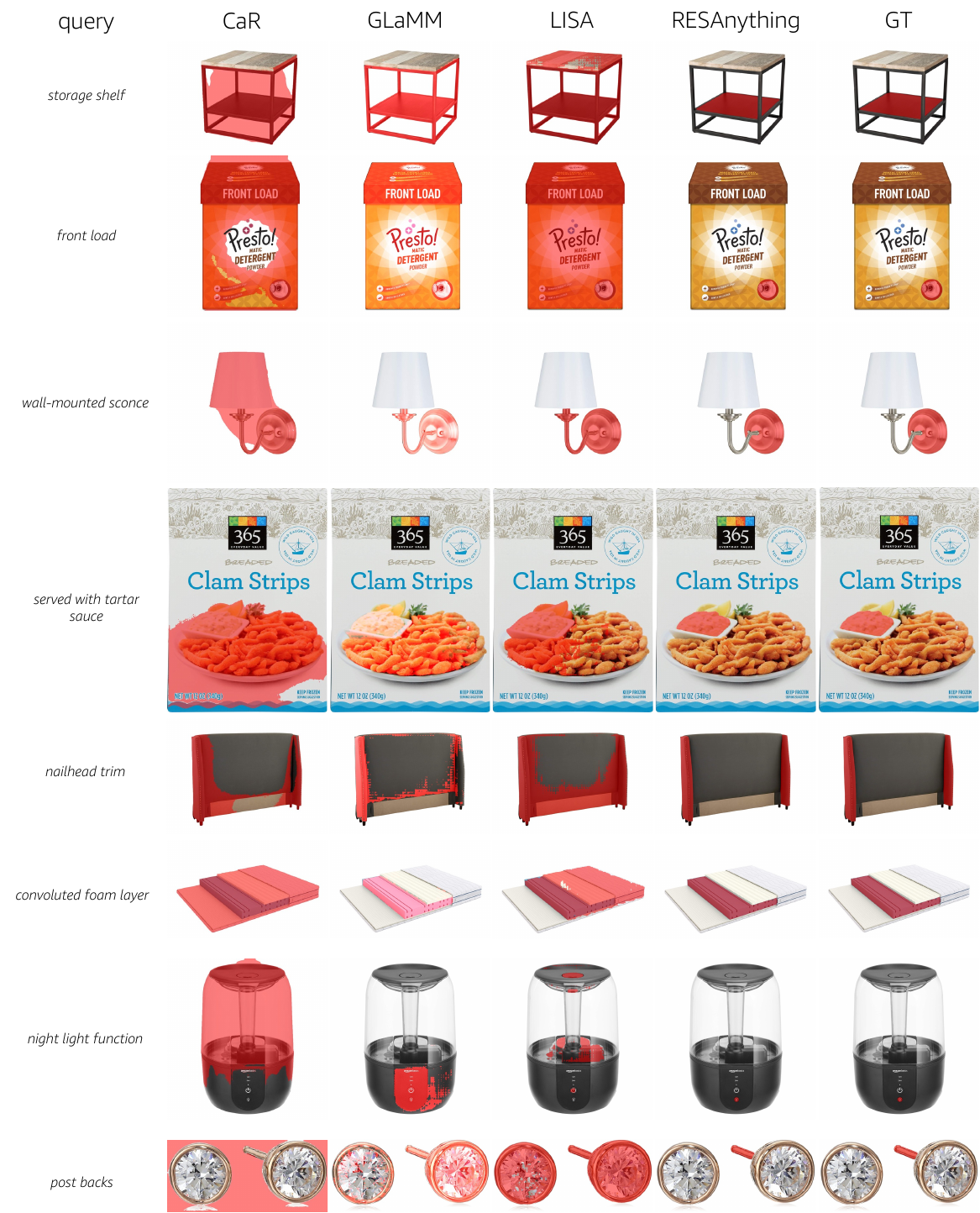}
    \caption{Qualitative results on \ourdataset~(Part 3). }
    \label{fig:qual-abo-3}
\end{figure*}

Lastly, we present qualitative comparisons on \ourdataset~(Figs.~\ref{fig:qual-abo-1}--\ref{fig:qual-abo-3}) to demonstrate our method's effectiveness on challenging expressions involving parts, materials, functionality, and package elements. See figure captions for detailed analysis.


\clearpage
{
    \small
    \bibliographystyle{ieeenat_fullname}
    \bibliography{main}
}
\end{document}